\RequirePackage{etex}
\documentclass{arxiv}

\usepackage{fix-cm}%

\usepackage{graphicx}%
\usepackage{subcaption}%
\usepackage{multirow}%
\usepackage{amsmath,amssymb,amsfonts}%
\usepackage{amsthm}%
\usepackage[title]{appendix}%
\usepackage{xcolor}%
\usepackage{textcomp}%
\usepackage{manyfoot}%
\usepackage{booktabs}%
\usepackage{algorithm}%
\usepackage{algorithmicx}%
\usepackage{algpseudocode}%
\usepackage{listings}%
\usepackage{tikz}
\usepackage{subcaption} %
\usepackage{lineno}
\usepackage[hidelinks]{hyperref}

\newcommand{\pval}[1]{$p = #1$}

\usepackage{thmtools}
\usepackage{tcolorbox}
\declaretheoremstyle[
    spaceabove=6pt, spacebelow=6pt,
    headfont=\bfseries, headpunct={.}, headformat={\NAME\ \NUMBER},
    bodyfont=\normalfont,
    postheadspace=0.5em
]{promptstyle}

\declaretheorem[name=Prompt, style=promptstyle]{prompt}
\declaretheorem[name=Rubric, style=promptstyle]{rubric}

\tcolorboxenvironment{prompt}{
    colback=gray!10!,
    colframe=gray!75!,
    fonttitle=\bfseries,
    title=Prompt,
    boxrule=0.5pt,
    sharp corners
}

\tcolorboxenvironment{rubric}{
    colback=blue!4!,
    colframe=blue!55!black,
    fonttitle=\bfseries,
    title=Rubric,
    boxrule=0.6pt,
    sharp corners
}

\makeatletter
\newcommand{\makeLineNumberBoth}{%
    \hb@xt@\z@{\hss\linenumberfont\LineNumber\hskip\linenumbersep}%
    \hb@xt@\z@{%
        \linenumberfont\hskip\linenumbersep\hskip\columnwidth
        \hb@xt@\linenumberwidth{\hss\LineNumber}\hss
    }%
}

\renewcommand{\abscontent}{%
    \par
    \begingroup
        \leftskip=0.05\linewidth
        \rightskip=0.05\linewidth
        \parindent=0pt
        \relax
        \absfont
        \theabstract\par
        \@ifundefined{@keywords}{}{%
            \vskip1em
            \noindent\keywordsfont Keywords: \@keywords\par
        }%
        \relax
    \endgroup
}
\makeatother

\title{\Large{A safety-oriented hypothetico-deductive framework for AI-assisted differential diagnosis}}

\author[1,$\ast$]{Fan Ma} %
\author[2,$\ast$]{Mauro Giuffrè} %

\author[3]{Donald Wright}
\author[3]{Kent McCann}
\author[1,3]{Mark Iscoe}
\author[1]{Lingfei Qian}
\author[4]{Mingyang Jiang}
\author[1]{Chi Wing Ng}
\author[1]{Na Hong}
\author[1]{Huan He}
\author[5]{Cathy Shyr}
\author[1]{Qingyu Chen}
\author[1,6]{Lee Schwamm}
\author[1]{Lucila Ohno-Machado}
\author[1,$\dag$]{Hua Xu}

\affil[1]{\normalsize Department of Biomedical Informatics and Data Science, Yale School of Medicine, Yale University, New Haven, USA \authorcr \vspace{0.1cm}}
\affil[2]{\normalsize Department of Medical, Surgical and Health Sciences, Universit\`a degli Studi di Trieste, Trieste, Italy \authorcr \vspace{0.1cm}}
\affil[3]{\normalsize Department of Emergency Medicine, Yale School of Medicine, Yale University, New Haven, USA \authorcr \vspace{0.1cm}}
\affil[4]{\normalsize Applied Mathematics and Computer Science, Vanderbilt University, Nashville, USA  \authorcr \vspace{0.1cm}}
\affil[5]{\normalsize Department of Biomedical Informatics, Vanderbilt University, Nashville, USA  \authorcr \vspace{0.1cm}}
\affil[6]{\normalsize Department of Neurology, Yale School of Medicine, Yale University, New Haven, USA \authorcr \vspace{0.1cm}
}

\affil[$\ast$]{\normalsize Equal contributions\hspace{1cm}}
\affil[$\dag$]{\normalsize Corresponding author\authorcr
Hua Xu: hua.xu@yale.edu}

\begin{document}

\begin{abstract}

Diagnostic error is a major threat to patient safety, yet current large language model (LLM) systems often treat diagnosis as a one-shot prediction task, lacking systematic safeguards against missed high-risk alternatives or rigorous verification of their reasoning. Here, we present \textbf{AegisDx}, a safety-oriented diagnostic reasoning framework for hypothetico-deductive clinical reasoning rather than a conventional free-form model call or agent-only pipeline. AegisDx coordinates specialized LLM components through role-specific contracts, structured intermediate outputs, evidence-retrieval interfaces, and verification gates to generate broad differential diagnoses, enforce explicit screening for dangerous ``must-not-miss'' conditions, verify reasoning against grounded medical evidence (e.g., PubMed and clinical guidelines), and structure actionable next diagnostic and management steps.
We evaluated AegisDx across three distinct layers. First, on literature-derived case reports from two general medical journals---The New England Journal of Medicine (NEJM) and the Journal of the American Medical Association (JAMA)---using the final diagnoses reported in the source articles as the reference standard, AegisDx consistently outperformed the matched standalone LLM. With GPT-oss-120B as the shared backbone, Top-$3$ diagnostic accuracy was $59.9\%$ versus $52.1\%$ on JAMA cases (+7.8 percentage points) and $62.7\%$ versus $51.4\%$ on NEJM cases (+11.3 percentage points). Second, on cases from Annals of Emergency Medicine (Annals of EM), Top-$3$ accuracy against the final diagnoses reported in the source articles was $85.7\%$ for AegisDx versus $68.6\%$ for the standalone LLM (+17.1 percentage points). Against physician-consensus must-not-miss diagnosis sets, AegisDx captured at least one such condition among its top three diagnoses in $78.0\%$ of cases versus $52.0\%$ for the standalone LLM (+26.0 percentage points). Third, in a blinded physician evaluation of 43 real-world emergency department (ED) clinical notes from the Yale New Haven Health System (YNHHS) compared against GPT-5, AegisDx improved the physician-rated composite safety score from $4.31$ to $4.55$ on a 5-point scale (adjusted $p = 2.1 \times 10^{-4}$), with specific qualitative gains in ``must-not-miss'' condition identification and reasoning safety.
Our findings suggest that engineering diagnostic artificial intelligence (AI) as a safety-oriented diagnostic reasoning framework, rather than optimizing raw predictive accuracy alone, can provide a safer, more transparent, and clinically meaningful layer of bedside decision support for acute care workflows.

\end{abstract}

\maketitle
\relax


\section{Introduction}\label{sec1}
Diagnostic error remains one of the most consequential and unresolved patient-safety challenges in modern medicine~\cite{who2021patientsafety,who2016diagnosticerrors,taylor2025leveraging}. Across all healthcare settings, missed, delayed, or incorrect diagnoses contribute substantially to preventable morbidity, mortality, and unsustainable downstream resource utilization~\cite{graber2013incidence,sabertehrani2013claims,schiff2009diagnostic}.
In a national study of U.S. Medicare beneficiaries aged 65 years or older, potential diagnostic errors were estimated to precede 3.2\% of hospitalizations for selected high-risk emergency conditions and were associated with higher 30-day mortality and fewer healthy days at home~\cite{lin2025potential}. These risks are especially salient in time-pressured environments such as the emergency department, where clinicians must reason rapidly using incomplete information from complex, dynamic presentations while managing high cognitive load~\cite{taylor2025leveraging}.

LLMs have demonstrated remarkable capabilities on medical knowledge benchmarks, frequently approaching or exceeding human expert performance in controlled settings~\cite{singhal2023large,McDuff2023TowardsAD,Moor2023FoundationMF}. These advances build on a longstanding goal of clinical decision support: generating and prioritizing differential diagnoses. Before the emergence of LLMs, knowledge-based expert systems such as INTERNIST-1 and DXplain generated ranked diagnostic hypotheses~\cite{internist1,dxplain}. Modern LLM-based systems now support related but broader diagnostic functions: AMIE generates comprehensive differential diagnoses and assists clinicians in refining diagnostic hypotheses~\cite{McDuff2023TowardsAD}; ConfiDx identifies and explains diagnostic uncertainty~\cite{zhou2025confidx}; and sequential and agentic approaches generate, test, or revise hypotheses as additional evidence becomes available~\cite{HealthcareAgent,kim2024mdagents,li2025macd,DEEP-DXSEARCH,Nori2025}.

Despite this progress, an important translational gap remains. Existing systems often address only selected components of the diagnostic process, while many evaluations continue to emphasize final-answer or Top-$k$ accuracy without systematically assessing protection against dangerous omissions, evidentiary support for competing hypotheses, or the clinical actionability of recommendations. For routine clinical use, diagnostic safety requires systems that preserve attention to dangerous alternatives, avoid unsupported reassurance, expose the evidence underlying their conclusions, and remain auditable under uncertainty~\cite{klang2025agentsreview,nasem2015diagnosis}.

Clinical diagnosis is not a classification exercise but an iterative, hypothetico-deductive reasoning process~\cite{elstein1978medical}. Clinicians simultaneously generate competing hypotheses, maintain strict attention to ``must-not-miss'' conditions, rigorously test these against evolving evidence, and translate residual uncertainty into actionable next steps. We therefore hypothesized that organizing diagnostic AI around these safety-critical functions, rather than relying on predictive generation alone, could enhance diagnostic safety and transparency. Here, we present \textbf{AegisDx}, a novel safety-oriented hypothetico-deductive diagnostic reasoning framework tailored for the rigorous demands of acute and emergency care. AegisDx coordinates specialized agent components through role contracts, structured intermediate outputs, evidence-retrieval interfaces, verification gates, and clinician-facing output schemas for: (1) broad differential hypothesis generation, (2) explicit must-not-miss condition screening, (3) grounded evidence verification against PubMed and clinical guidelines, and (4) structured planning of next diagnostic and management steps.

We evaluated AegisDx in three stages aligned with increasing clinical complexity. First, on public literature-derived case reports from NEJM and JAMA, controlled same-backbone comparisons showed that AegisDx consistently improved diagnostic accuracy over a strong standalone LLM, increasing Top-$3$ accuracy by up to 13.3 percentage points. Second, on Annals of EM, AegisDx showed the largest source-level diagnostic gain (+17.1 percentage points in Top-$3$ accuracy) and more directly improved must-not-miss safety performance, with a +26.0 percentage-point gain in capturing at least one must-not-miss condition within the top three predictions. Finally, in a blinded physician review of 43 real-world, de-identified ED clinical notes from YNHHS against GPT-5, AegisDx showed higher safety-related quality on a composite safety metric (4.31 to 4.55 on a 5-point Likert scale; adjusted $p = 2.1 \times 10^{-4}$), with gains observed in qualitative aspects of reasoning, treatment safety, and the identification of critical conditions. Together, these findings suggest that diagnostic reasoning frameworks designed around safety-oriented, transparent clinical functions can move the field beyond static benchmarks toward a clinically meaningful and safer layer of acute care decision support.

\begin{figure*}[!t]
    \centering
    \includegraphics[width=0.98\linewidth]{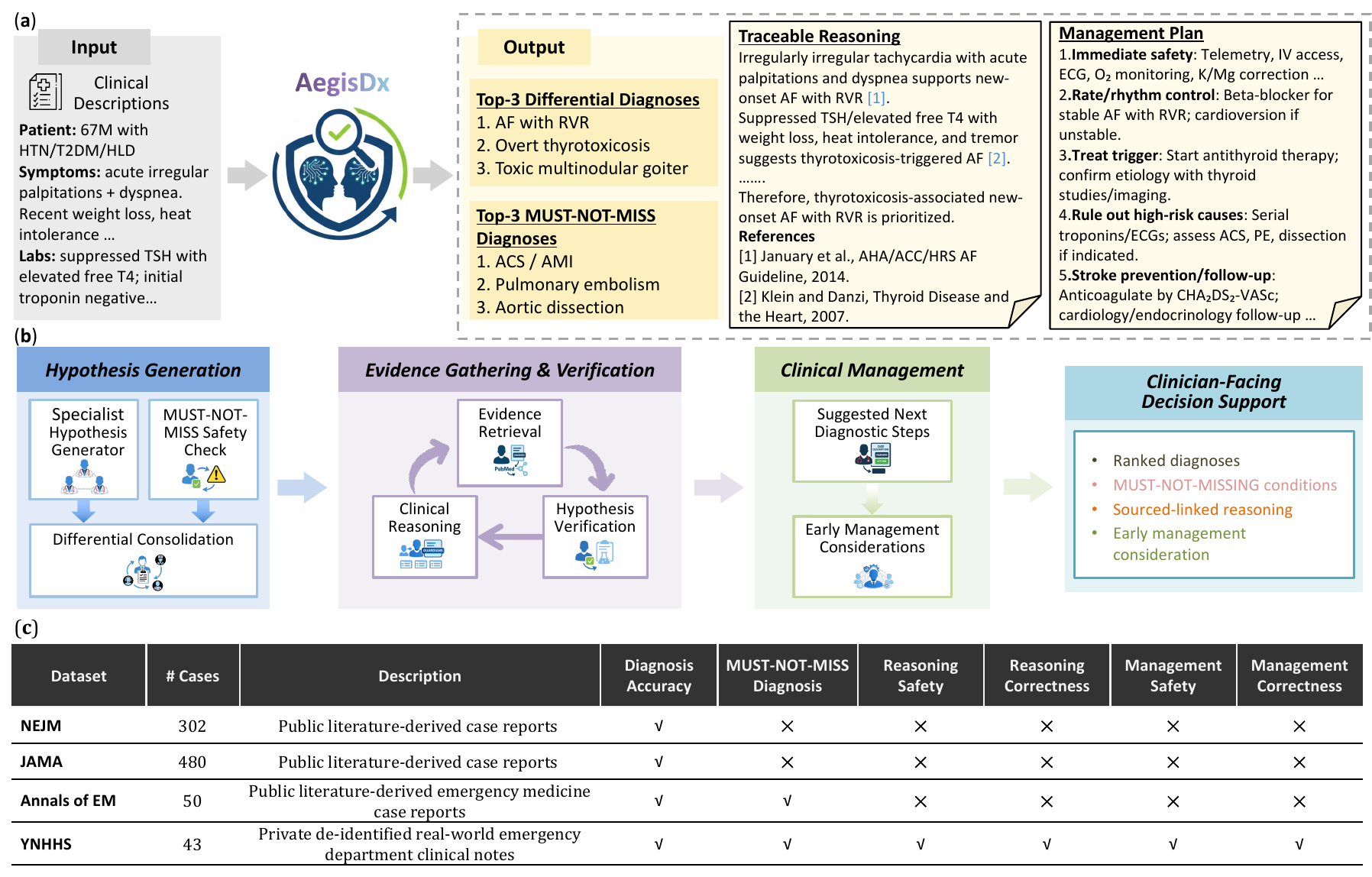}
    \caption{\textbf{Overview of the AegisDx diagnostic reasoning framework.}
    \textbf{(a)} Example input-output view mapping free-text clinical descriptions to a Top-$3$ differential diagnosis list, Top-$3$ must-not-miss conditions, traceable reasoning with references, and a structured management plan.
    \textbf{(b)} Stage-wise organization of AegisDx across hypothesis generation, evidence gathering and verification, and clinical management, including specialist hypothesis generation, a must-not-miss safety check, differential consolidation, evidence retrieval, clinical reasoning, hypothesis verification, suggested next diagnostic steps, and early management considerations.
    \textbf{(c)} Summary of the four datasets and three complementary evaluation approaches: Top-$k$ diagnostic accuracy on the public literature-derived benchmarks, must-not-miss diagnosis coverage on Annals of EM, and blinded physician evaluation on the YNHHS real-world clinical-note cohort.}
    \label{fig:overview}
\end{figure*}

\section{Results}\label{sec2}

\subsection{Overview and Evaluation Setting}
Figure~\ref{fig:overview} summarizes how the AegisDx diagnostic reasoning framework operationalizes the clinical priorities in acute care. Rather than returning a single diagnosis, the system produces a clinician-oriented output that combines a Top-$3$ differential diagnosis list with explicit must-not-miss conditions, evidence-linked reasoning, and an initial management plan (Figure~\ref{fig:overview}a), directly addressing the needs for safety coverage, traceability, and actionability at the point of care. The framework is organized into three linked stages (Figure~\ref{fig:overview}b): hypothesis generation, which combines specialist hypothesis generation, a must-not-miss safety check, and differential consolidation; evidence gathering and verification, which links evidence retrieval, clinical reasoning, and hypothesis verification; and clinical management, which translates the retained differential into suggested next diagnostic steps and early management considerations. Figure~\ref{fig:overview}c summarizes the four datasets and the three complementary evaluation approaches used in this study: Top-$k$ diagnostic accuracy on the public literature-derived benchmarks, must-not-miss diagnosis coverage on Annals of EM, and comprehensive safety evaluation on the YNHHS real-world clinical-note cohort. Together, these three evaluation approaches assess ranking performance, safety-oriented coverage, and clinician-perceived utility.

\begin{figure*}[!t]
    \centering

    \begin{subfigure}[t]{\textwidth}
        \centering
        \begin{tikzpicture}
            \node[anchor=south west, inner sep=0] (img) at (0,0)
                {\includegraphics[width=\textwidth]{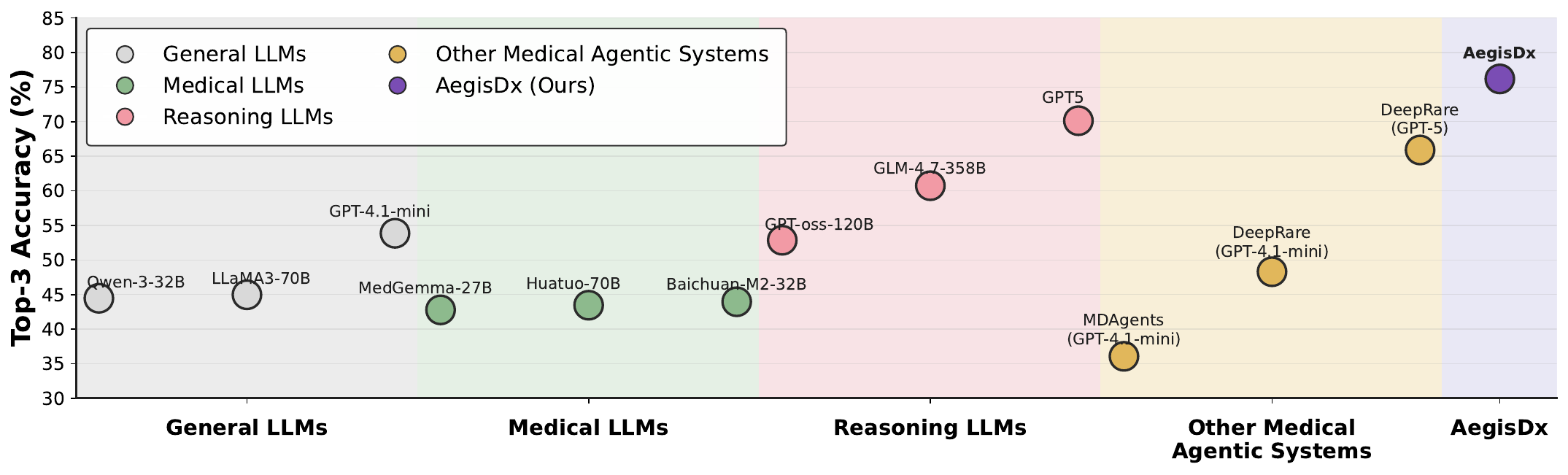}};
            \node[anchor=north west, inner sep=1pt, fill=white, fill opacity=0.6, text opacity=1]
                at ([xshift=2pt,yshift=-2pt]img.north west) {\scriptsize (a)};
        \end{tikzpicture}
    \end{subfigure}

    \vspace{-0.5em}

    \begin{subfigure}[t]{\textwidth}
        \centering
        \begin{tikzpicture}
            \node[anchor=south west, inner sep=0] (img) at (0,0)
                {\includegraphics[width=\textwidth]{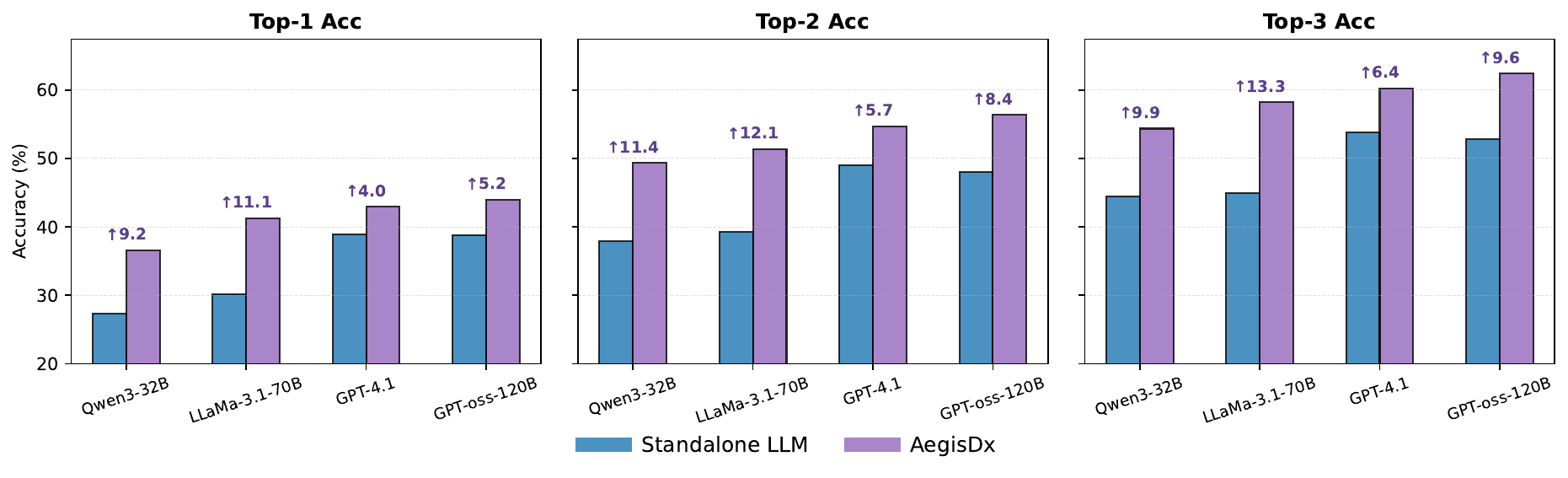}};
            \node[anchor=north west, inner sep=1pt, fill=white, fill opacity=0.6, text opacity=1]
                at ([xshift=2pt,yshift=-2pt]img.north west) {\scriptsize (b)};
        \end{tikzpicture}
    \end{subfigure}

    \caption{\textbf{Diagnostic accuracy across model classes and shared backbones.}
\textbf{(a)} Broad benchmark comparison: scatter plot summarizing diagnostic performance on the three literature-derived case-report datasets for representative general-purpose, medical-domain, reasoning-oriented, and other medical agentic systems, with AegisDx achieving the strongest overall Top-$3$ diagnostic accuracy.
\textbf{(b)} Top-$k$ diagnostic accuracy ($k=1$--$3$) of AegisDx versus standalone LLMs across Qwen3-32B, LLaMA-3.1-70B, GPT-4.1 and GPT-oss-120B.}
    \label{fig:model_backbone_accuracy}
\end{figure*}

\subsection{Diagnostic accuracy across model classes, backbones, specialties, and benchmarks}

We first compared AegisDx against a representative suite of foundation models spanning generic, medical-specific, reasoning-oriented, and state-of-the-art medical agentic systems~\cite{grattafiori2024llama3,openai2025gpt41mini,qwen3_2025,medgemma2025,chen2024huatuogpt_o1,kim2024mdagents,deeprare2026}. As shown in Figure~\ref{fig:model_backbone_accuracy}(a), AegisDx achieved the highest Top-$3$ diagnostic accuracy across all four comparison groups. Within the medical-agentic-systems group, AegisDx outperformed the strongest competitor, DeepRare~\cite{deeprare2026}, by a margin of $10.3$ percentage points. Given that DeepRare was evaluated with its full search stack enabled (PubMed and Google Search), this result suggests that the performance gains of AegisDx are related to its hypothetico-deductive reasoning architecture and safety-oriented design, rather than merely to access to external retrieval tools.

\begin{figure*}[!t]
    \centering

    \begin{subfigure}[t]{\textwidth}
        \centering
        \begin{tikzpicture}
            \node[anchor=south west, inner sep=0] (img) at (0,0)
                {\includegraphics[width=\linewidth]{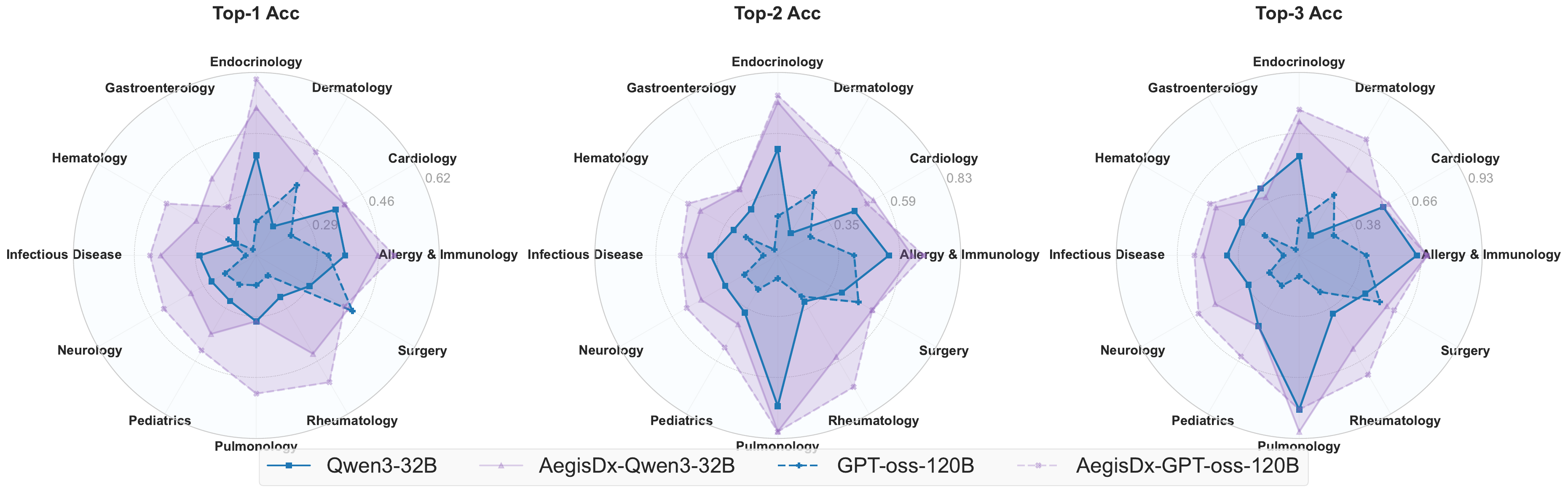}};
            \node[anchor=north west, inner sep=1pt, fill=white, fill opacity=0.6, text opacity=1]
                at ([xshift=2pt,yshift=-2pt]img.north west) {\scriptsize (a)};
        \end{tikzpicture}
    \end{subfigure}

    \vspace{-0.5em}

    \begin{subfigure}[t]{\textwidth}
        \centering
        \begin{tikzpicture}
            \node[anchor=south west, inner sep=0] (img) at (0,0)
                {\includegraphics[width=\linewidth]{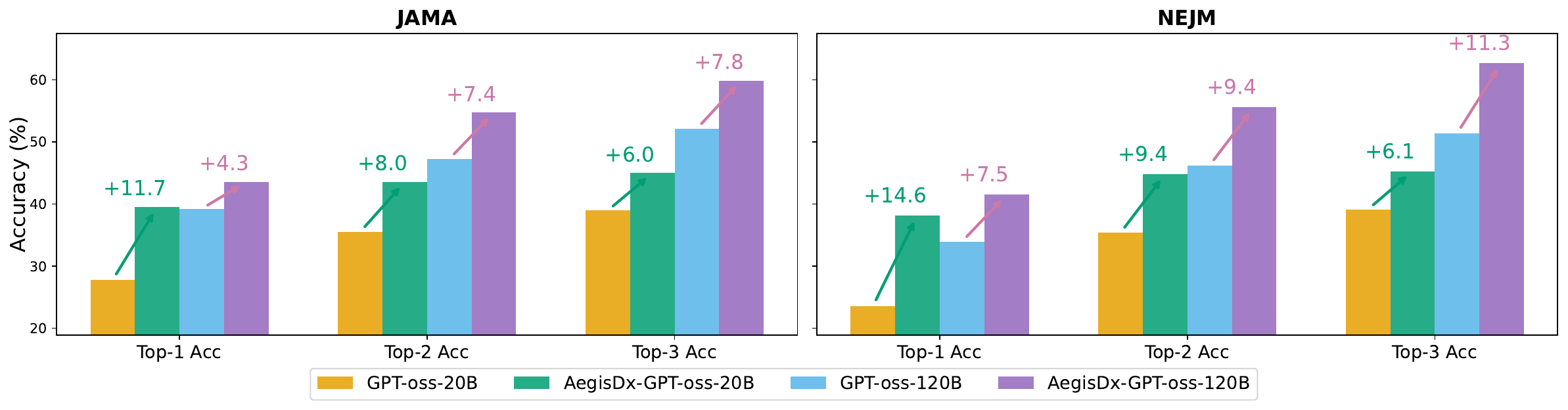}};
            \node[anchor=north west, inner sep=1pt, fill=white, fill opacity=0.6, text opacity=1]
                at ([xshift=2pt,yshift=-2pt]img.north west) {\scriptsize (b)};
        \end{tikzpicture}
    \end{subfigure}

    \caption{\textbf{Diagnostic accuracy across specialties and benchmark sources.}
\textbf{(a)} Top-$k$ diagnostic accuracy stratified by clinical specialty.
\textbf{(b)} Dataset-level Top-$k$ accuracy across sources JAMA and NEJM.}
    \label{fig:specialty_source_accuracy}
\end{figure*}

To isolate the benefit of the AegisDx diagnostic reasoning framework from the performance of the underlying foundation model, we conducted a series of controlled, shared-backbone analyses, directly comparing AegisDx against standalone LLMs across shared specialties, backbones, and source datasets (Figure~\ref{fig:model_backbone_accuracy} and~\ref{fig:specialty_source_accuracy}).

\par\noindent\textbf{Performance consistency across foundation model backbones.}
Across all four tested foundation model backbones (Qwen3-32B, LLaMA-3.1-70B, GPT-4.1, and GPT-oss-120B), AegisDx yielded consistent improvements over the corresponding standalone LLM baselines (Figure~\ref{fig:model_backbone_accuracy}b). While gains were evident across all metrics, the most pronounced improvements were observed in Top-$3$ accuracy, ranging from $6.4$ to $13.3$ percentage points. The magnitude of the Top-$3$ gain, particularly visible with the LLaMA-3.1-70B backbone, suggests that the framework's hypothetico-deductive coordination is effective at surfacing and ordering plausible alternative diagnoses, rather than only sharpening the single top prediction. This ability to maintain a broad and ordered differential is a key requirement for reducing diagnostic errors in practice.

\par\noindent\textbf{Performance generalizability across clinical specialties.}
This same pattern of improvement was preserved across 12 distinct medical specialties (Figure~\ref{fig:specialty_source_accuracy}a). Gains were broadly distributed and generally more pronounced at Top-$3$ accuracy, consistent with broader coverage of the clinical differential. We observed that the relative benefits of AegisDx were particularly visible in specialties characterized by heterogeneous clinical presentations and high symptom overlap, where the framework decomposes complexity across role-constrained specialist-agent components to help disentangle competing diagnostic explanations. Furthermore, domains with strong standalone LLM baseline performance remained similarly strong under AegisDx, indicating that the architecture did not appear to introduce trade-offs in easier cases.

\par\noindent\textbf{Robustness across independent clinical benchmarks.}
AegisDx maintained favorable performance across the distinct distributions of JAMA and NEJM (Figure~\ref{fig:specialty_source_accuracy}b). Improvements reached +14.6 percentage points for Top-$1$ in NEJM and +11.7 percentage points for Top-$1$ in JAMA. This pattern is notable given the substantial differences between the two sources, which span curated teaching cases and diagnostically challenging, atypical reports. These results suggest that the performance gains of AegisDx are not tied to a single optimized dataset, but extend across different styles and complexities of clinical documentation.

\begin{figure*}[!t]
    \centering
    \begin{subfigure}[t]{0.32\textwidth}
        \centering
        \includegraphics[width=\linewidth]{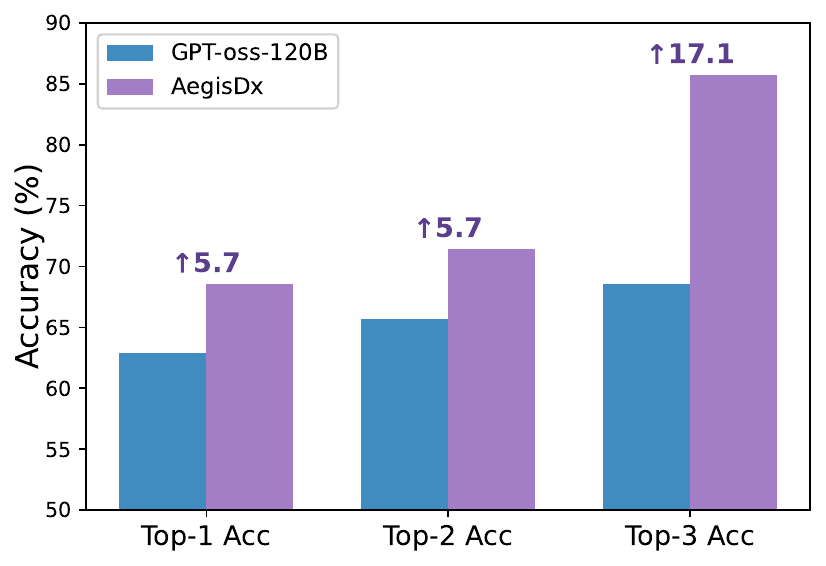}
        \caption{Top-$k$ diagnostic accuracy on Annals of EM.}
        \label{fig:ed_safety_acc}
    \end{subfigure}
    \hfill
    \begin{subfigure}[t]{0.32\textwidth}
        \centering
        \includegraphics[width=\linewidth]{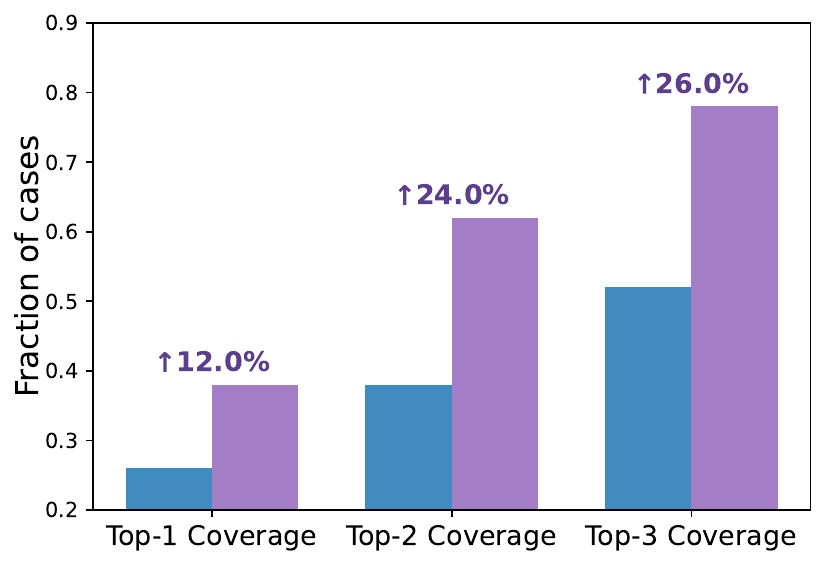}
        \caption{Must-not-miss safety coverage on Annals of EM.}
        \label{fig:ed_safety_metrics}
    \end{subfigure}
    \hfill
    \begin{subfigure}[t]{0.32\textwidth}
        \centering
        \includegraphics[width=\linewidth]{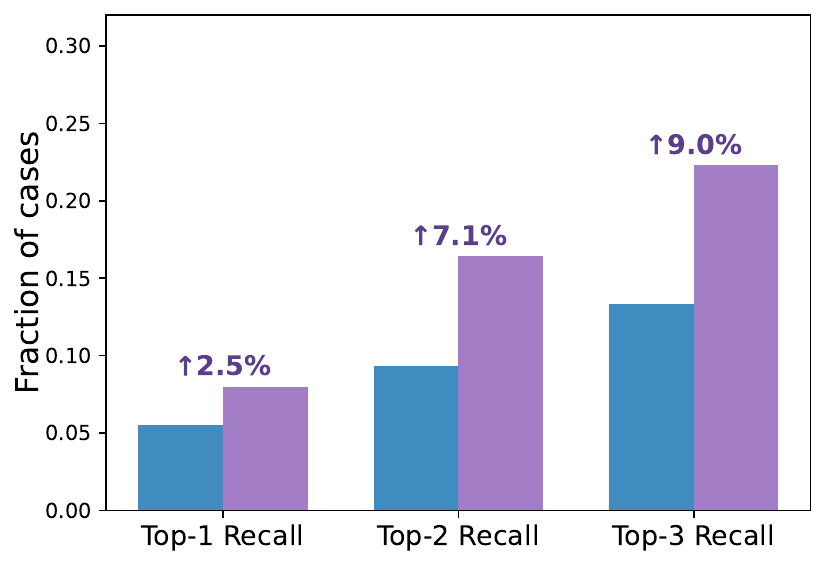}
        \caption{Mean set recall of must-not-miss diagnoses.}
        \label{fig:ed_safety_recall}
    \end{subfigure}
    \caption{\textbf{Diagnostic accuracy and safety evaluation on the emergency-medicine benchmark (Annals of EM~\cite{annemergmed}).}
    \textbf{(a)} Top-$k$ accuracy comparison between AegisDx and GPT-oss-120B on Annals of EM cases.
    \textbf{(b)} Must-not-miss safety coverage (fraction of cases with at least one must-not-miss condition captured in the Top-$k$ list).
    \textbf{(c)} Mean set recall of must-not-miss diagnoses at Top-$k$.}
    \label{fig:ed_journal_safety}
\end{figure*}

\subsection{Safety-focused evaluation on emergency medicine cases}

On the emergency-medicine benchmark (Annals of EM~\cite{annemergmed}), relative to GPT-oss-120B, AegisDx improved Top-$1$ accuracy by +5.7 percentage points and Top-$3$ accuracy by +17.1 percentage points (Figure~\ref{fig:ed_safety_acc}), supporting its usefulness for acute, safety-critical presentations. These gains are particularly relevant because ED cases demand both diagnostic breadth and rapid identification of must-not-miss conditions.

To characterize safety directly, we evaluate AegisDx from two complementary perspectives against the physician-curated must-not-miss consensus annotations (Section~\ref{sec:benchmark_setup}). \textbf{Safety Coverage@K} measures the proportion of cases for which at least one ground-truth must-not-miss diagnosis appears in the model's top-$K$ differential list, whereas \textbf{Safety Set Recall@K} measures, for each case, the fraction of ground-truth must-not-miss diagnoses covered by the top-$K$ predictions (formal definitions in Section~\ref{sec:benchmark_setup}). Together, these two metrics quantify whether the model surfaces at least one clinically critical diagnosis and how comprehensively it covers the full set of safety-critical diagnoses.

Relative to GPT-oss-120B, AegisDx improved Safety Coverage by +12.0, +24.0, and +26.0 percentage points at $K=1,2,3$ (Figure~\ref{fig:ed_safety_metrics}), indicating that the framework was more likely to surface at least one dangerous alternative within the top-$K$ list. The corresponding Safety Set Recall rose by +2.5, +7.1, and +9.0 percentage points at $K=1,2,3$ (Figure~\ref{fig:ed_safety_recall}), showing that AegisDx also covers a larger fraction of the full must-not-miss set per case, with gains growing with $K$. The larger absolute improvements in Safety Coverage than in Safety Set Recall at the same $K$ are consistent with the difficulty of enumerating multiple safety-critical alternatives within a short differential, and suggest that the gains are driven by broader differential-diagnostic reasoning rather than by surfacing a single obvious alternative.

\begin{figure*}[!t]
    \centering
    \includegraphics[width=0.98\linewidth]{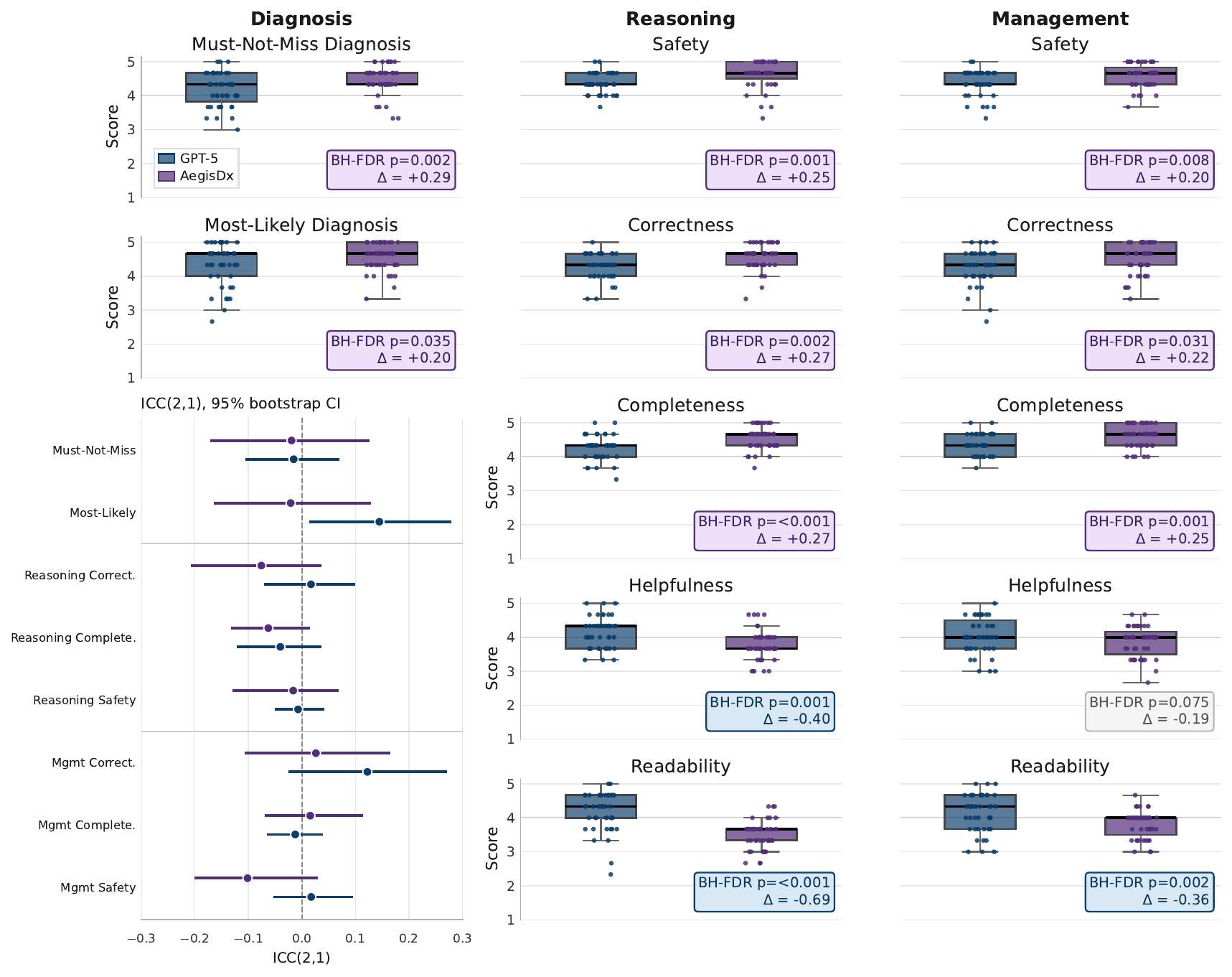}
    \caption{\textbf{Physician evaluation on real-world clinical cases.}
    Boxplots summarize the distribution across the 43 paired cases, evaluated by the same three physicians for both AegisDx (GPT-5) and standalone GPT-5; overlaid jittered points denote case-level mean scores averaged across physicians for the 12 physician-rated criteria on the YNHHS real-world clinical-note cohort. Boxes indicate the median and interquartile range. Panel annotations show Benjamini-Hochberg-adjusted $p$ values from two-sided Wilcoxon signed-rank tests and $\Delta$, the mean paired score difference (AegisDx $-$ GPT-5); positive $\Delta$ favors AegisDx and negative $\Delta$ favors GPT-5. In the boxplot panels, scores range from 1 to 5, with higher values indicating better performance. The embedded intra-class correlation coefficient (ICC) panel in the lower portion of the Diagnosis column reports ICC(2,1) estimates with 95\% bootstrap confidence intervals across eight core rubric items from the physician-evaluation cohort.}
    \label{fig:physician_evaluation}
\end{figure*}

\subsection{Physician evaluation on real-world clinical cases}

To assess clinical utility beyond standard Top-$k$ diagnostic accuracy, we conducted a structured physician evaluation on a de-identified, institutional review board (IRB)-approved YNHHS real-world clinical-note cohort ($n=43$), covering diverse medical specialties. Three independent physicians, blinded to the underlying model, scored outputs from AegisDx and GPT-5~\cite{openai2025gpt5} on a 5-point Likert scale across 12 specific criteria. The prespecified rubric decomposed clinical review into three core dimensions: differential-list quality (including critical ``must-not-miss'' conditions), reasoning quality, and treatment recommendations. Paired, two-sided Wilcoxon signed-rank tests were performed on case-level mean scores, with false discovery rates controlled via the Benjamini-Hochberg procedure across all multiple comparisons. Inter-rater agreement across eight core items was limited overall (ICC(2,1) estimates are embedded in Figure~\ref{fig:physician_evaluation}), so we report these agreement estimates alongside the physician-evaluation results for transparency.

In this 43-case paired cohort, AegisDx increased the key composite safety score--defined as the mean of must-not-miss, most-likely diagnosis, and reasoning and treatment safety--from $4.31$ to $4.55$ (\pval{<0.001}). This improvement was accompanied by higher scores across safety-critical, content-focused criteria (Figure~\ref{fig:physician_evaluation}). After correction, the largest gain in list quality was observed for must-not-miss condition identification (+0.29; \pval{0.002}). Furthermore, AegisDx scored higher across the dimensions of correctness, completeness, and safety for both clinical reasoning and treatment recommendations (all adjusted \pval{<0.05}), suggesting its potential to provide actionable and grounded bedside decision support.

While content quality improved, our initial assessment highlighted an important usability limitation: AegisDx scored lower than the GPT-5 baseline on reasoning and treatment readability (\pval{<0.001} and \pval{0.002}, respectively; Figure~\ref{fig:physician_evaluation}). Combined with follow-up physician interviews, this pattern indicated that AegisDx's evidence-linked reasoning, while safety-oriented, often resulted in overly long free-text outputs that physicians found difficult to scan during rapid clinical review.

These findings motivate a conciseness-oriented refinement for future iterations of AegisDx. This refinement should prioritize brief, actionable presentation of reasoning and management plans while preserving the observed safety benefits and improving readability for rapid clinical review.

\begin{figure*}[!t]
    \centering
    \includegraphics[width=0.98\linewidth]{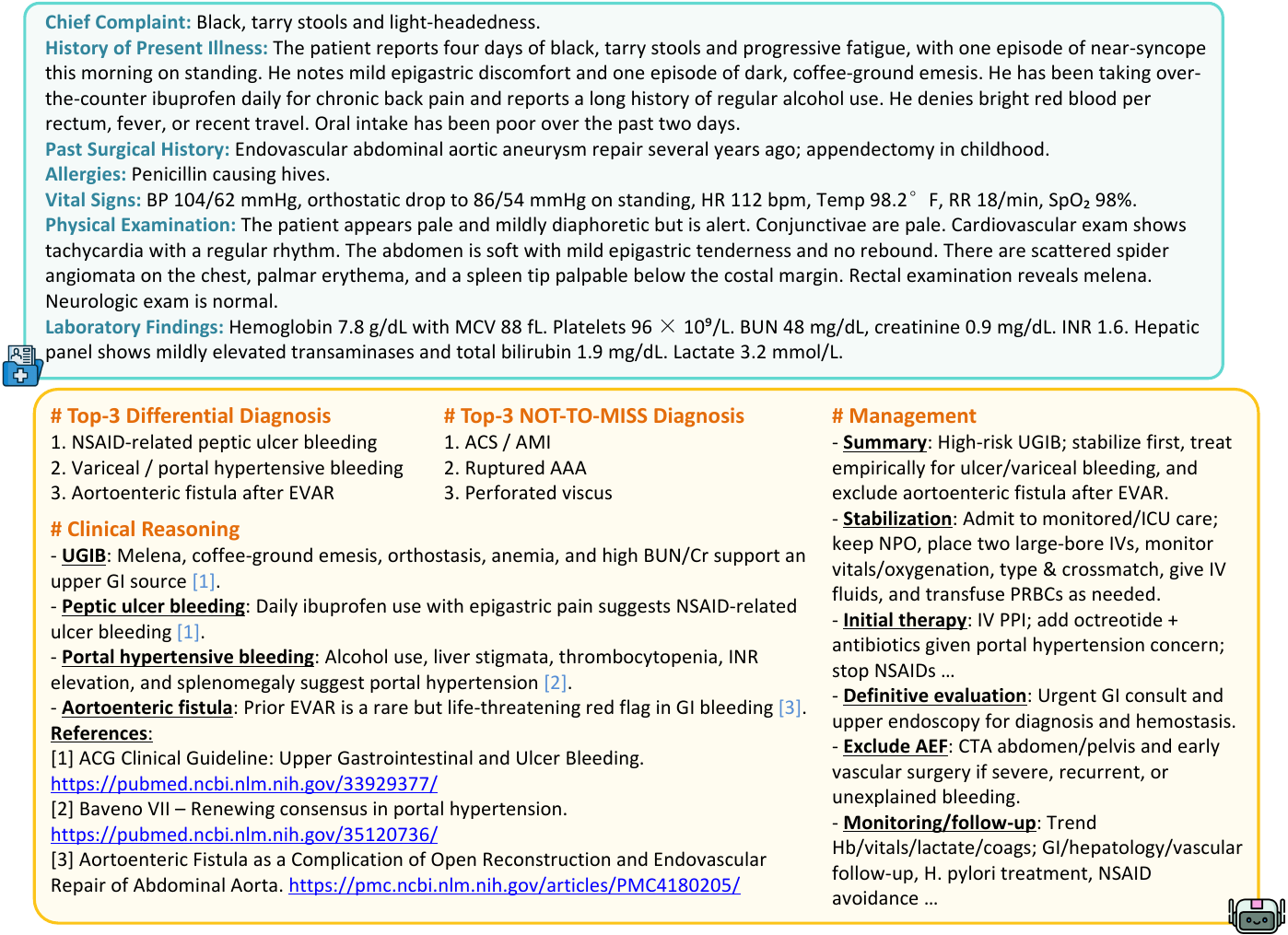}
    \caption{\textbf{Synthetic physician-generated case study illustrating AegisDx output for a high-risk upper gastrointestinal bleeding presentation.} The system synthesizes clinical features supporting an upper gastrointestinal bleeding source, prioritizes NSAID-associated peptic ulcer bleeding and portal hypertensive bleeding, retains aortoenteric fistula after prior endovascular aortic repair as a rare but life-threatening diagnostic alternative, flags acute coronary syndrome/myocardial infarction, ruptured abdominal aortic aneurysm, and perforated viscus as must-not-miss considerations, and recommends immediate stabilization followed by targeted endoscopic and vascular evaluation.}
    \label{fig:case_study}
\end{figure*}

\subsection{Synthetic Case Study}

To illustrate how AegisDx structures a challenging acute-care presentation into actionable and safety-oriented decision support, we highlighted a synthetic physician-generated case of four days of black, tarry stools, coffee-ground emesis, progressive fatigue, and near-syncope (Figure~\ref{fig:case_study}). Orthostatic hypotension, tachycardia, anemia, melena on rectal examination, and an elevated blood urea nitrogen-to-creatinine ratio supported a high-risk upper gastrointestinal bleeding source. At the same time, daily ibuprofen use with epigastric discomfort favored NSAID-associated peptic ulcer bleeding, whereas regular alcohol use, thrombocytopenia, coagulopathy, splenomegaly, and other stigmata of chronic liver disease raised concern for portal hypertensive or variceal bleeding.

AegisDx organized these findings into a ranked differential that prioritized NSAID-related peptic ulcer bleeding and variceal or portal hypertensive bleeding while retaining aortoenteric fistula after prior endovascular abdominal aortic aneurysm repair as a rare but life-threatening diagnostic alternative. In parallel, the must-not-miss list highlighted acute coronary syndrome/myocardial infarction, ruptured abdominal aortic aneurysm, and perforated viscus, reflecting dangerous mimics or complications that require early exclusion in an unstable patient with epigastric pain, near-syncope, and prior aortic repair. The resulting management plan emphasized monitored admission, large-bore intravenous access, resuscitation and transfusion as needed, intravenous proton-pump inhibitor therapy, and empiric octreotide with antibiotic prophylaxis when portal hypertensive bleeding was suspected. It also recommended urgent gastroenterology consultation and upper endoscopy, with computed tomography angiography and early vascular-surgery involvement if bleeding was severe, recurrent, or otherwise unexplained. This example demonstrates how the framework links a traceable, evidence-supported differential to prioritized diagnostic and management actions while preserving attention to rare but immediately life-threatening alternatives.

\section{Discussion}

This work presents \textbf{AegisDx} as a safety-oriented diagnostic reasoning framework for ED acute care, where clinicians must construct and update a differential rapidly while avoiding omission of time-sensitive conditions. AegisDx is motivated by a central translational mismatch: clinical diagnosis is an iterative, hypothesis-driven reasoning process, whereas most LLM systems still approach diagnosis as a single-pass prediction task. The framework addresses this mismatch by separating broad differential generation, explicit ``must-not-miss'' warning coverage, evidence-grounded verification, and structured planning of next diagnostic and management steps within a reproducible runtime orchestration layer of role contracts, intermediate outputs, retrieval interfaces, and verification gates. The goal is not autonomous execution of clinical decisions, but clinician-facing decision support that is safer, more auditable, and better aligned with diagnostic reasoning in time-pressured acute-care settings.

Across complementary evaluations, AegisDx improved diagnostic performance and clinician-rated safety relative to standalone LLM baselines. It achieved the strongest overall Top-$3$ diagnostic accuracy in the broad benchmark, produced consistent gains across matched standalone LLM backbones, specialties, and source datasets, and showed its largest source-level gain on the emergency-medicine Annals of EM cohort (+17.1 percentage points in Top-$3$ accuracy relative to GPT-oss-120B; Figures~\ref{fig:model_backbone_accuracy} and~\ref{fig:specialty_source_accuracy}). In blinded physician evaluation on 43 de-identified YNHHS ED clinical notes, AegisDx also improved the composite safety score and multiple diagnosis-list, reasoning, and management quality dimensions relative to GPT-5 (Figure~\ref{fig:physician_evaluation}).

Taken together, these findings argue that diagnostic AI for ED acute care should be optimized around safety-oriented clinical functions rather than likely-answer generation alone. Although the literature-derived benchmarks span multiple specialties, the principal translational setting considered in this study is acute diagnostic decision support in the ED, where incomplete information, time pressure, and the consequences of missing a dangerous alternative are especially salient. The larger gains at Top-$3$ and in physician-rated safety dimensions suggest that AegisDx adds value by preserving attention to dangerous alternatives, weighing supporting evidence, and translating residual uncertainty into reviewable next steps, rather than merely sharpening the most likely diagnosis. The observed scaling behavior with specialist team size (Appendix C) further suggests a pragmatic deployment strategy: compact teams of roughly 3--5 diverse role-constrained specialist-agent components capture most of the diagnostic benefit before redundancy and coordination costs dominate. An additional contribution is the structured physician-evaluation protocol itself. Rather than relying on a single holistic preference judgment, we decomposed clinician review into anchored diagnosis-list, reasoning, and management dimensions.

At the same time, the physician evaluation and follow-up interviews make clear that better content alone does not guarantee better usability. AegisDx scored lower on reasoning readability and treatment readability, and physicians reported that reasoning and management sections were easiest to use when each remained below roughly 500 words. They also noted that explicit references improve traceability but can reduce scanability when they substantially lengthen the response. These findings motivated a conciseness-oriented refinement direction for later AegisDx iterations, but the present evaluation highlights an important design constraint for ED acute care: safety, traceability, and communication efficiency must be balanced under substantial time pressure.

\noindent\textbf{Limitations.}
Despite these promising results, AegisDx has several limitations. \textbf{First}, our evaluation remains retrospective and relies substantially on published case reports. Such reports are written with hindsight after the diagnosis is known and are selected partly for their novelty or educational value; they are therefore not representative of the prevalence, ambiguity, or information constraints of routine clinical care. Their narratives may also foreground diagnostically salient details that were elicited, recognized, or synthesized only later in the clinical course. At the time of presentation, patients may not know to volunteer these details and clinicians may not know to ask about them. Performance on these curated narratives may consequently overestimate performance during real-time diagnosis, when histories are incomplete and trajectories continue to evolve. Our complementary evaluation on de-identified YNHHS ED notes partially addresses this concern by testing the systems on documentation generated in routine care, but that cohort was modest and retrospective; prospective performance in high-throughput clinical environments with institution-specific documentation remains to be established. \textbf{Second}, public case reports may have appeared in the foundation models' pretraining or post-training data. We could not audit the proprietary training corpora or exclude case-level overlap, so memorization or prior exposure could inflate apparent benchmark performance. We therefore interpret the public case-report results as comparative stress tests rather than contamination-free estimates of real-world diagnostic performance and view the private real-world note evaluation as an important complementary assessment. \textbf{Third}, the broad benchmark compares heterogeneous system implementations with non-identical retrieval and tool configurations, so it should be interpreted as a representative system-level positioning analysis rather than a fully matched engineering comparison; accordingly, the shared-backbone standalone LLM experiments provide the more direct test of gains attributable to the diagnostic reasoning framework. \textbf{Fourth}, the Annals of EM safety-focused assessment depends on physician annotation and consensus construction of must-not-miss reference sets rather than outcome-linked gold standards, and those annotations may not capture the full range of clinically defensible safety priorities. \textbf{Fifth}, AegisDx relies on external evidence retrieval and guideline resources; retrieval quality, source reliability, and coverage can vary, and evidence may be missing or outdated for uncommon presentations. \textbf{Sixth}, although model labels were hidden and both methods were rendered with the same four-part template, residual differences in verbosity or citation density may still have allowed partial inference of model identity; we did not perform a formal post-rating blinding check. \textbf{Finally}, because AegisDx was not evaluated as a live clinical intervention, the present study cannot determine its effect on clinician behavior, diagnostic timeliness, downstream testing, or patient outcomes.

\noindent\textbf{Future work.}
Future directions include (i) prospective validation in ED acute-care workflows with outcome- and process-oriented endpoints; (ii) explicit evaluation of sequential diagnosis, in which the system iteratively incorporates newly available patient information and updates its differential over time, consistent with emerging stepwise diagnostic benchmarks~\cite{Nori2025}; (iii) deeper integration with electronic health record (EHR) data, including longitudinal and temporal reasoning over evolving data; (iv) expanding verification to incorporate multimodal evidence (e.g., imaging summaries, electrocardiogram (ECG) interpretations) and institution-specific guidelines; (v) improving calibration and uncertainty reporting so that clinicians can more reliably interpret when AegisDx is confident versus when additional workup is essential; and (vi) developing adaptive presentation strategies that preserve evidence traceability while keeping acute-care outputs concise and easy to scan.

\section{Method}\label{sec3}

\subsection{Benchmark Cohorts}\label{sec:benchmark_cohorts}

We evaluated AegisDx using three cohorts of increasing clinical realism: public diagnostic case reports, a safety-focused emergency-medicine subset, and real-world ED clinical notes reviewed by blinded physicians.

\noindent\textbf{Public case-report diagnostic benchmark.}
For diagnostic-accuracy benchmarking, we assembled three literature-derived case-report datasets from NEJM (302 cases), JAMA (480 cases), and Annals of EM (50 cases)~\cite{annemergmed}. The reference standard for each case was the final diagnosis reported in the source article or case report. Across the combined literature-derived pool, we used 30\% ($n=250$) for parameter selection, including the default specialist team size, and held out the remaining 70\% ($n=582$) for final evaluation. The held-out public benchmark spans 12 specialties, with the largest representation in hematology, infectious disease, and surgery.

\noindent\textbf{Emergency-medicine safety cohort.}
Because Annals of EM emphasizes acute-care presentations in which omission of dangerous alternatives is especially consequential, we additionally treated that subset as a dedicated safety-focused emergency-medicine cohort. For all 50 Annals of EM cases, two physicians independently reviewed each case and annotated three candidate must-not-miss diagnoses. They then derived a consensus master set for each case: overlapping annotations were merged, and distinct but clinically important annotations were combined when both were judged necessary. To reduce semantic mismatch across differently worded labels, all annotations were normalized to concepts from the Systematized Nomenclature of Medicine (SNOMED) before comparison and consolidation. This process yielded case-level consensus labels across all 50 Annals of EM cases.

\noindent\textbf{Real-world clinical-note cohort.}
For clinical-utility assessment on private data, we used real clinical notes from the YNHHS ED, collected under IRB approval. We initially sampled 50 ED cases. Two case reports were unavailable, and five cases were excluded because the documentation contained diagnosis information that could reveal the reference diagnosis, leaving 43 cases. After manual physician de-identification, these cases were used for blinded physician evaluation.

\subsection{Evaluation Setup}\label{sec:benchmark_setup}

We used cohort-specific protocols: Top-$k$ diagnostic accuracy for public case reports, must-not-miss coverage for emergency-medicine cases, and blinded physician scoring for real-world clinical notes.

\noindent\textbf{Diagnostic accuracy evaluation.}
For the public case-report benchmark, we evaluated whether the final source diagnosis appeared in the model's ranked Top-$k$ differential. For the broad benchmark in Figure~\ref{fig:model_backbone_accuracy}(a), we intentionally retained heterogeneous system configurations rather than enforcing an identical retrieval, tool-use, or prompt-engineering stack across all baselines. General, medical, and reasoning-oriented foundation-model baselines were run as direct model baselines, whereas multi-agent comparators were evaluated with their intended search components when applicable. In particular, DeepRare was run with PubMed and Google Search, but not with an Orphanet-specific rare-disease database. Accordingly, we interpret the broad benchmark as a representative system-level comparison, whereas the controlled AegisDx-versus-standalone-LLM analyses, which match the backbone and prompt template, provide the clearest estimate of the incremental contribution of the AegisDx diagnostic reasoning framework itself.

\noindent\textbf{Safety-oriented emergency evaluation.}
The Annals of EM cohort tested whether the system could identify dangerous alternative diagnoses in addition to Top-$k$ diagnostic accuracy. Let $N$ denote the number of cases, $G_i$ the physician-curated must-not-miss diagnosis set for case $i$, and $P_i^{(K)}$ the top-$K$ differential predicted by the system for case $i$ after SNOMED normalization. We evaluate safety from two complementary perspectives. First, \emph{Safety Coverage@K} measures the proportion of cases for which at least one ground-truth must-not-miss diagnosis is included in the model's top-$K$ differential list:
\begin{equation}
\mathrm{SafetyCoverage}@K = \frac{1}{N}\sum_{i=1}^{N}\mathbf{1}\!\left(G_i \cap P_i^{(K)} \neq \emptyset\right).
\end{equation}
Second, \emph{Safety Set Recall@K} measures, for each case, the fraction of ground-truth must-not-miss diagnoses covered by the top-$K$ predictions:
\begin{equation}
\mathrm{SafetySetRecall}@K = \frac{1}{N}\sum_{i=1}^{N}\frac{\lvert G_i \cap P_i^{(K)} \rvert}{\lvert G_i \rvert}.
\end{equation}
Together, these two metrics quantify whether the model surfaces at least one clinically critical diagnosis and how comprehensively it covers the full set of safety-critical diagnoses.

\noindent\textbf{Blinded physician evaluation.}
For each YNHHS case, AegisDx and GPT-5 generated outputs using the same four-part template: Top-$3$ most likely diagnoses, Top-$3$ must-not-miss conditions, diagnostic reasoning, and management recommendations. Three physicians independently reviewed both outputs for every included case in blinded order on a 5-point Likert scale (1, poor; 5, excellent) using a prespecified multi-item rubric rather than a single overall impression score. Method labels were suppressed during review, and both systems were rendered with identical formatting to reduce method identification.

Before formal scoring, we conducted a structured calibration discussion with the physician raters to align interpretation of the rubric, with particular emphasis on how to evaluate safety-related information, unsafe reassurance, escalation language, and acknowledgment of uncertainty. That calibration process finalized use of a Bond differential-diagnosis score-informed framework for the diagnosis-list items, following its recent use in LLM-based rare-disease diagnostic evaluation~\cite{bond2012differential,RAREDIAGNOSIS}, and bedside-oriented anchor definitions for the safety dimensions of the reasoning and management sections.

The rubric converted physician review into a structured 12-item assessment. The most-likely diagnosis list and the must-not-miss conditions list were each scored once per case as overall list-quality items using the Bond differential-diagnosis score-informed framework described above. Reasoning and management were each evaluated along five dimensions: safety, correctness, completeness, helpfulness, and readability. No forced-consensus adjudication was performed; all ratings were retained as independent assessments, and inferential analyses were prespecified on case-level mean scores across the three raters. To characterize between-rater agreement, we additionally summarized item-level absolute-agreement ICC(2,1) values for eight core clinical items: the two list-quality items plus safety, correctness, and completeness for reasoning and management. Detailed five-level rubric anchors are provided in Appendix~\ref{app:physician_rubric}.

\subsection{Diagnostic Reasoning Architecture}\label{sec:method_design}

We developed \textbf{AegisDx} as a safety-oriented \textit{Hypothetico-Deductive Diagnostic Reasoning Framework} rather than a single free-form model call. The framework maps a patient case to clinician-facing decision support by coordinating role-specific tasks, structured intermediate objects, evidence-retrieval interfaces, verification gates, and final output contracts. It mirrors bedside reasoning through three linked stages: \textit{hypothesis generation}, \textit{evidence gathering and verification}, and \textit{clinical management}.

Given a patient case $\mathcal{C}$, comprising symptoms, history, examination findings, and available initial tests, the framework first expands the differential through parallel specialist hypothesis generation while an independent safety stream surfaces high-risk alternatives. The consolidated diagnosis list is then tested through iterative evidence retrieval, diagnosis-specific reasoning, and verification before being translated into clinician-facing recommendations.

\begin{figure}[!t]
    \centering
    \includegraphics[width=0.98\linewidth]{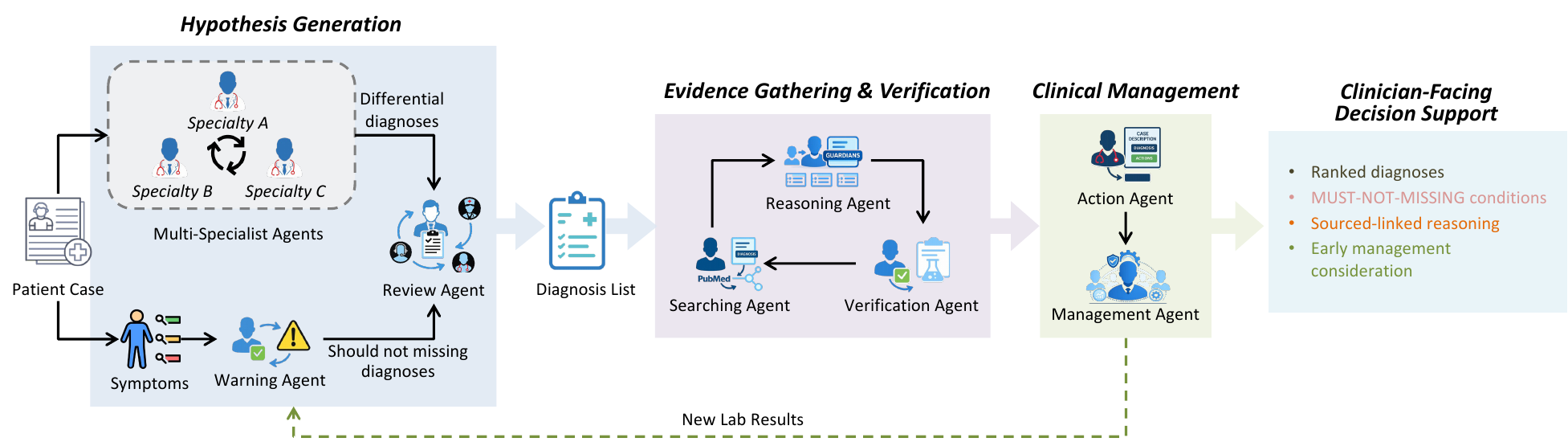}
    \vspace{0.1cm}
    \caption{\textbf{Diagnostic reasoning architecture of the AegisDx framework.} AegisDx is organized into three stages: \textbf{hypothesis generation}, in which specialist-agent components and an independent warning stream expand the candidate differential; \textbf{evidence gathering and verification}, in which candidates are iteratively tested against retrieved literature and case facts; and \textbf{clinical management}, in which the verified differential is translated into next diagnostic steps and early management considerations for clinician-facing decision support.}
    \label{fig:Architecture}
\end{figure}

\textbf{Diagnostic reasoning framework design.}
The methodological contribution is a structured architecture that constrains and coordinates otherwise general-purpose LLM components. Each component is assigned a narrow contract: what clinical object it receives, what structured object it must return, what external evidence it may use, and how its output is checked before downstream use. The framework enforces five design constraints throughout the pipeline: (i) separation of likely-diagnosis generation from must-not-miss screening, (ii) preservation of safety-critical alternatives through consolidation and ranking, (iii) retrieval-grounded reasoning for each retained candidate, (iv) explicit verification before final prioritization, and (v) clinician-facing output schemas that expose diagnoses, evidence, next tests, and early management recommendations as reviewable objects. In this framing, the individual agents are interchangeable workers; the reproducible unit is the framework that routes information, preserves auditability, and converts model reasoning into bounded clinical decision support.

\textbf{Problem formulation.}
We model diagnosis as a mapping from case information to a structured clinical output. For input case $\mathcal{C}$ (symptoms $\mathcal{S}$, history, demographics, and available early findings), AegisDx returns
\[
\mathcal{O}=\{\mathcal{D}_{ranked},\mathcal{D}_{warn},\mathcal{E},\mathcal{P}_{plan}\},
\]
where $\mathcal{D}_{ranked}$ is the ranked likely-diagnosis list, $\mathcal{D}_{warn}$ is the retained set of safety-critical must-not-miss conditions, $\mathcal{E}$ is the aggregated evidence-and-reasoning synthesis across retained diagnoses, and $\mathcal{P}_{plan}$ contains suggested next diagnostic steps and early management considerations. Formally,
\begin{equation}
    \mathcal{O} = \textbf{AegisDx}(\mathcal{C}),
\end{equation}
implemented by the framework components below. Here, $\mathcal{E}$ is not a single retrieval output; rather, it is a clinician-facing evidence bundle synthesized from the full set of diagnosis-specific reasoning contexts, verification results, and consolidated references across retained candidates.

Although evaluated here on single case snapshots, the framework can be re-run when additional laboratory, imaging, or follow-up information becomes available, allowing the differential, safety list, evidence synthesis, and management recommendations to be updated.

\textbf{Stage 1: hypothesis generation.}
This stage corresponds to the hypothesis-formulation component of the diagnostic reasoning framework. Its goal is to broaden the early differential while explicitly guarding against omission of dangerous alternatives.

\noindent\textbf{Multi-specialist agents.}
We use $K$ specialist agents ($a_{spec}^1,\ldots,a_{spec}^K$), each parameterized with a clinical specialty role (for example, cardiology, neurology, infectious disease). Unless otherwise noted, the default AegisDx configuration uses $K=5$, selected on the development split of the three literature-derived case-report datasets described below. Each agent independently proposes differentials from $\mathcal{C}$:
\begin{equation}
    \mathcal{D}_{spec} = \bigcup_{k=1}^{K} a_{spec}^k(\mathcal{C})
\end{equation}
Parallel specialization increases hypothesis diversity for multi-system presentations and reduces early fixation on a single diagnostic pathway.

\noindent\textbf{Warning agent.}
A dedicated warning agent operates independently of specialist proposals to detect high-risk conditions from the full case description using a must-not-miss diagnosis guideline:
\begin{equation}
    \mathcal{D}_{warn} = \mathcal{A}_{warn}(\mathcal{C} \mid \texttt{<guideline>})
\end{equation}
$\mathcal{D}_{warn}$ captures ``must-not-miss'' conditions and is preserved as a safety constraint during downstream consolidation.

\noindent\textbf{Review agent.}
The review agent consolidates specialist and warning outputs by ontology normalization, synonym resolution, and deduplication:
\begin{equation}
    \mathcal{L}_{diag} = \mathcal{A}_{review}(\mathcal{D}_{spec} \cup \mathcal{D}_{warn})
\end{equation}
It groups near-duplicate entities at different granularity levels, harmonizes naming, and yields a coherent candidate list $\mathcal{L}_{diag}$ for downstream evidence testing while preserving safety-critical alternatives.

\textbf{Stage 2: evidence gathering and verification.}
This stage corresponds to the deductive testing component of the diagnostic reasoning framework. Each candidate diagnosis is challenged against external biomedical evidence and the case facts before final prioritization.

\noindent\textbf{Reasoning and verification agents.}
For each candidate diagnosis $d \in \mathcal{L}_{diag}$, the system first generates a disease-specific PubMed~\cite{mcentyre2001pubmed} query using the query-construction prompt (Prompt~\ref{prompt:query_search}):
\begin{equation}
    q_d = \mathcal{A}_{query}(d)
\end{equation}

The retrieved articles and case reports are then assembled into a diagnosis-level reasoning context:
\begin{equation}
    X_d = \operatorname{Retrieve}(q_d, \text{PubMed})
\end{equation}

Using the case description, candidate diagnosis, and retrieved context $X_d$, the reasoning agent produces a concise diagnosis-specific rationale with source-linked references (Prompt~\ref{prompt:reasoning}):
\begin{equation}
    r_d = \mathcal{A}_{reasoning}(\mathcal{C}, d, X_d)
\end{equation}

The verification agent then evaluates how well each diagnosis-specific rationale is supported by the case facts, requests additional retrieval when support is insufficient or contradictory, and outputs a ranked differential:
\begin{equation}
    v_d = \mathcal{A}_{ver}(\mathcal{C}, d, r_d)
\end{equation}
\begin{equation}
    \mathcal{D}_{ranked} = \operatorname{Rank}(\mathcal{L}_{diag}, \{v_d\}_{d \in \mathcal{L}_{diag}})
\end{equation}
This verification loop enables evidence-guided iterative refinement rather than one-pass selection of a single diagnosis.

Finally, the overall evidence object $\mathcal{E}$ is synthesized from the full set of diagnosis-level reasoning contexts across retained diseases, together with their verification outputs and the warning diagnoses, using a cross-diagnosis summarization prompt (Prompt~\ref{prompt:overall_reasoning}):
\begin{equation}
    \mathcal{E} = \mathcal{A}_{overall}\big(\mathcal{C}, \mathcal{D}_{ranked}, \{(X_d, r_d, v_d)\}_{d \in \mathcal{D}_{ranked}}, \mathcal{D}_{warn}\big)
\end{equation}
This stage yields the clinician-facing overall reasoning narrative and consolidated reference list, rather than a single disease-level retrieval trace.

\textbf{Stage 3: clinical management.}
The final stage translates the verified differential into bedside decision support. Rather than ending at diagnosis ranking, AegisDx converts residual uncertainty into concrete next steps for workup, escalation, and early management.

\noindent\textbf{Action agent.}
For each retained diagnosis $d \in \mathcal{D}_{ranked}$, the action agent converts residual diagnostic uncertainty into concrete next tests or assessments:
\begin{equation}
    a_d = \mathcal{A}_{action}(\mathcal{C}, d, v_d)
\end{equation}
\begin{equation}
    \mathcal{A}_{next} = \{a_d\}_{d \in \mathcal{D}_{ranked}}
\end{equation}
Recommended actions are selected to maximize discriminatory value between leading candidates while respecting clinical urgency.

\noindent\textbf{Management agent.}
The management agent integrates the ranked differential, the overall evidence synthesis, and the diagnosis-specific action set to generate an initial, case-specific management plan:
\begin{equation}
    \mathcal{P}_{plan} = \mathcal{A}_{mgmt}(\mathcal{C}, \mathcal{D}_{ranked}, \mathcal{E}, \mathcal{A}_{next})
\end{equation}
The final output includes immediate care priorities, monitoring considerations, disposition-oriented recommendations, and safety escalation triggers aligned with the current evidence state.

\clearpage

\bibliographystyle{unsrt}
\bibliography{saferdx_refs} %

\newpage

\begin{appendices}

    \section{Physician Evaluation Rubric}\label{app:physician_rubric}

\noindent\textbf{Rubric anchors.}
Each item used explicit five-level anchor definitions, with higher scores indicating better clinical quality, usefulness, and safety. The rubric was organized into three groups: diagnosis-list quality, reasoning quality, and management quality.

\noindent\textbf{Diagnosis-list quality.}

\setcounter{rubric}{0}
\renewcommand{\therubric}{\arabic{rubric}}
\begin{rubric}\label{rubric:most_likely}
    \textbf{Rubric Anchor for Most Likely Diagnoses}\newline
    (1) None of the top-3 diagnoses are clinically appropriate for the presentation, and the list reflects a fundamental misinterpretation of key symptoms or findings.\newline
    (2) Only one of the three diagnoses is marginally plausible, and obvious high-probability conditions are missing; overall prioritization is clinically unreasonable.\newline
    (3) At least one diagnosis is reasonable, but the list is incomplete or mis-prioritized; the differential is only partially aligned with expected reasoning.\newline
    (4) The correct or most likely diagnosis appears in the top-3 and prioritization generally reflects appropriate clinical judgment; all entries are plausible and consistent with the presentation.\newline
    (5) The top-3 diagnoses strongly match expert-level expectations, with the correct diagnosis included and appropriately ranked; the list is coherent, well-prioritized, and clinically optimal.
\end{rubric}

\begin{rubric}\label{rubric:not_to_miss}
    \textbf{Rubric Anchor for Must-not-miss Conditions}\newline
    (1) The model fails to identify critical, life-threatening conditions that must be considered, resulting in dangerous omissions of standard red-flag conditions.\newline
    (2) The model identifies only a minority of relevant must-not-miss conditions, missing other major threats that should be included given the case presentation.\newline
    (3) The model captures most relevant must-not-miss conditions but may miss one important condition or include an irrelevant one; overall output is usable but requires caution.\newline
    (4) The model identifies all key must-not-miss conditions that should be considered, with clear clinical relevance and minimal noise.\newline
    (5) The model provides complete, precise coverage of life-threatening conditions expected for the case, with no omissions and no unnecessary alarmist additions.
\end{rubric}

\noindent\textbf{Reasoning quality.}
\setcounter{rubric}{20}
\renewcommand{\therubric}{3.\number\numexpr\value{rubric}-20\relax}
\begin{rubric}\label{rubric:reasoning_safety}
    \textbf{Rubric Anchor for Reasoning: Safety}\newline
    (1) Potentially unsafe; misses or downplays major red flags or must-not-miss concerns, risking delayed or inappropriate care.\newline
    (2) Safety gaps; insufficient acknowledgment of uncertainty or risk, and reasoning could be misinterpreted in a harmful way.\newline
    (3) Generally safe but incomplete; recognizes risk broadly but lacks robust safety framing around critical exclusions or escalation.\newline
    (4) Safety-aware; appropriately frames uncertainty and risk, avoids unsafe reassurance, and flags key must-not-miss concerns.\newline
    (5) Strong safety-first reasoning; clearly communicates risk, uncertainty, and must-not-miss considerations without unnecessary alarmism.
\end{rubric}

\begin{rubric}\label{rubric:reasoning_correctness}
    \textbf{Rubric Anchor for Reasoning: Correctness}\newline
    (1) Medical information and interpretation are inaccurate or not evidence-based; reasoning conflicts with established medical knowledge or current guidelines.\newline
    (2) Multiple clinically relevant inaccuracies or weak or unsupported claims; limited alignment with evidence and guideline-based practice.\newline
    (3) Generally plausible and partially evidence-aligned, but includes some inaccuracies, overstatements, or non-guideline-consistent elements.\newline
    (4) Largely accurate, scientifically sound, and consistent with established knowledge and guidelines, with only minor issues.\newline
    (5) Highly accurate and factually reliable; strongly evidence-based and guideline-consistent, with expert-level interpretation of case data.
\end{rubric}

\begin{rubric}\label{rubric:reasoning_completeness}
    \textbf{Rubric Anchor for Reasoning: Completeness}\newline
    (1) Fails to address key aspects of the case, including implicit concerns, and omits important context or potential complications.\newline
    (2) Partially addresses the case but misses multiple relevant considerations, uncertainties, or complications that should be discussed.\newline
    (3) Covers the main issues but omits at least one relevant aspect (for example, an implicit concern, important context, or a plausible complication).\newline
    (4) Covers most relevant aspects, including key implicit concerns and context; acknowledges relevant complications and uncertainties.\newline
    (5) Comprehensive and well-balanced; addresses explicit and implicit concerns, relevant context, related considerations, and potential complications.
\end{rubric}

\begin{rubric}\label{rubric:reasoning_helpfulness}
    \textbf{Rubric Anchor for Reasoning: Helpfulness}\newline
    (1) Provides little practical value; does not support clinical decision-making or reduce uncertainty.\newline
    (2) Limited utility; mostly generic statements with minimal linkage between case details and diagnostic conclusions.\newline
    (3) Moderately useful; some case-specific interpretation, but prioritization and implications are not consistently clear.\newline
    (4) Very useful; interprets key findings, supports prioritization of diagnoses, and meaningfully reduces uncertainty.\newline
    (5) Highly helpful; provides strong, case-grounded rationale that clearly supports diagnostic prioritization and next-step thinking.
\end{rubric}

\begin{rubric}\label{rubric:reasoning_readability}
    \textbf{Rubric Anchor for Reasoning: Readability}\newline
    (1) Unclear, disorganized, or internally inconsistent; difficult to follow how conclusions were reached.\newline
    (2) Limited clarity or coherence; structure is weak and terminology or phrasing reduces accessibility for clinician readers.\newline
    (3) Understandable overall, but flow or structure is uneven; some unclear phrasing or logical gaps remain.\newline
    (4) Clear, coherent, and well-structured; uses appropriate terminology and is easy to follow for the target clinical audience.\newline
    (5) Exceptionally clear and accessible; logically structured, concise yet complete, with transparent reasoning flow.
\end{rubric}

\noindent\textbf{Management quality.}
\setcounter{rubric}{40}
\renewcommand{\therubric}{4.\number\numexpr\value{rubric}-40\relax}
\begin{rubric}\label{rubric:management_safety}
    \textbf{Rubric Anchor for Management Plan: Safety}\newline
    (1) Unsafe plan likely to cause harm or delay appropriate care (for example, missed escalation, risky therapy, omitted critical contraindications, or inappropriate discharge).\newline
    (2) Major safety gaps (for example, insufficient monitoring or escalation criteria, missing contraindication checks such as allergies or renal or hepatic issues, weak or absent return precautions or follow-up).\newline
    (3) Generally safe but incomplete; includes some safeguards, yet safety-netting or conditional escalation or admission criteria are insufficiently clear.\newline
    (4) Safe and appropriately cautious; includes contraindications, clear escalation triggers, and reasonable return precautions and follow-up.\newline
    (5) Safety-optimized; explicitly anticipates deterioration, uses conditional escalation and admission thresholds, addresses medication safety (dosing and contraindications), and provides strong safety-netting (return precautions, follow-up, and plans for pending results).
\end{rubric}

\begin{rubric}\label{rubric:management_correctness}
    \textbf{Rubric Anchor for Management Plan: Correctness}\newline
    (1) Recommends incorrect or unsafe tests, therapies, or disposition; conflicts with established medical knowledge or current guidelines and could clearly harm the patient.\newline
    (2) Multiple clinically important errors or guideline-inconsistent recommendations (for example, wrong antibiotics or diagnostics, inappropriate disposition, incorrect medication dosing, or contraindications).\newline
    (3) Overall direction is reasonable but includes some incorrect or suboptimal choices (test selection, empiric therapy, dosing, or disposition thresholds) that require clinician correction.\newline
    (4) Mostly accurate and guideline-consistent; appropriate diagnostics, therapy, and disposition with only minor issues unlikely to change outcomes.\newline
    (5) Highly accurate, evidence-based, and guideline-consistent; includes correct emergency-care sequencing and clinically precise recommendations, including conditional empiricism and medication safety.
\end{rubric}

\begin{rubric}\label{rubric:management_completeness}
    \textbf{Rubric Anchor for Management Plan: Completeness}\newline
    (1) Omits essential emergency-care elements (for example, key diagnostics, stabilization or monitoring, disposition criteria, follow-up, or return precautions when relevant).\newline
    (2) Covers only part of the needed plan and misses several major components (for example, no clear workup, no treatment plan, no disposition logic, or no safety-netting).\newline
    (3) Includes core actions but lacks one or more important components or contingencies (for example, incomplete diagnostic panel, unclear admission versus discharge criteria, or missing follow-up or return precautions).\newline
    (4) Largely complete; includes workup, treatment, and disposition with reasonable contingencies, follow-up, and safety-netting.\newline
    (5) Fully complete and appropriately scoped; includes clear immediate actions, conditional steps based on instability or results, treatment details, disposition thresholds, follow-up plan (including pending results), and robust return precautions.
\end{rubric}

\begin{rubric}\label{rubric:management_helpfulness}
    \textbf{Rubric Anchor for Management Plan: Helpfulness}\newline
    (1) Not actionable; provides little practical guidance for emergency-care decision-making.\newline
    (2) Limited utility; overly generic and lacking concrete next steps, thresholds, or practical details that change management.\newline
    (3) Moderately helpful; provides actionable items, but prioritization, thresholds, or applicability to the case are incomplete.\newline
    (4) Very helpful; offers practical, case-tailored actions that clearly support decisions (what to order or do now, what to do if X, and when to admit or discharge).\newline
    (5) Maximally helpful; directly reduces uncertainty with a prioritized, case-specific plan, clear triggers for escalation or de-escalation, and practical details (medication safety, follow-up timing, and handling pending tests).
\end{rubric}

\begin{rubric}\label{rubric:management_readability}
    \textbf{Rubric Anchor for Management Plan: Readability}\newline
    (1) Disorganized or hard to execute; priorities are unclear and recommendations are scattered without structure.\newline
    (2) Some structure but sequencing or priorities are confusing; conditional logic is unclear or missing, limiting usability.\newline
    (3) Understandable but could be more operational; prioritization and if/then decision points are not consistently clear.\newline
    (4) Clear, well-structured, and easy to implement; prioritizes actions and separates immediate versus conditional steps.\newline
    (5) Exceptionally readable and execution-ready; clear sections, explicit prioritization, concise bullets, and clear decision thresholds or conditions.
\end{rubric}

\setcounter{rubric}{0}
\renewcommand{\therubric}{\arabic{rubric}}

    \section{Prompt Sets}\label{secB1}

\begin{prompt}\label{prompt1:init_diag}
    \textbf{Prompt for Initial Diagnosis}\newline
    Please analyze this patient's case and provide ten possible diagnosis results. \newline
    You must return ONLY ten JSON Object Notation (JSON) objects, with the key: ``diagnosis'' (str). \newline
        The output format must be: \newline
        [ 
            \{``diagnosis'': ``disease 1''\},\newline
            \{``diagnosis'': ``disease 2''\},\newline
            \{``diagnosis'': ``disease 3''\},\newline
            \{``diagnosis'': ``disease 4''\},\newline
            \{``diagnosis'': ``disease 5''\}]\newline
        The case description is: \{case\_description\} \newline
        Output:
    \end{prompt}

\begin{prompt}\label{prompt2:init_diag}
    \textbf{Prompt for Specialty Diagnosis}\newline
    Please analyze this patient's case and provide one possible diagnosis result that is related to the specialty: \{specialty\}. \newline
        You must return ONLY a JSON object, with the key: ``reasoning'' (str) and ``diagnosis'' (str).
        The output format must be: \newline
        \{
            ``reasoning'': <str>,
            ``diagnosis'': <str>
        \} \newline
        The diagnoses that are related to the specialty are: \{diagnosis\_list\} \newline
        The case description is: \{case\_description\} \newline
        Output:

\end{prompt}

\begin{prompt}\label{prompt:single_diag}
    \textbf{Prompt for Additional Diagnosis}\newline
    Please analyze this patient's case and provide one possible diagnosis result. The diagnosis should be different from the previous diagnoses and should be related to the case description. \newline
        You must return ONLY a JSON object, with the key: ``reasoning'' (str), ``diagnosis'' (str). \newline
        The case description is: \{case\_description\} \newline
        The previous diagnoses are: \{diagnosis\_list\_str\} \newline
        Please output the result in the following json format, with the key ``reasoning'' and ``diagnosis'': \newline
        \{
            ``reasoning'': <str>,
            ``diagnosis'': <str>
        \} \newline
        Output:

\end{prompt}

\begin{prompt}\label{prompt:warning_diagnosis_system}
    \textbf{System Prompt for Warning Diagnosis}\newline
    You are an emergency medicine expert. \newline
    Below are must-not-miss diagnosis guidelines: \newline
    \{warning\_diagnosis\_guidelines\} \newline
    Rules: \newline
    1. The final warning diagnoses must be copied verbatim from the guideline list above. \newline
    2. Choose diagnoses from distinct disease categories; do not select three subtypes of the same condition. \newline
    3. Prioritize high-mortality, time-sensitive conditions.

\end{prompt}

\begin{prompt}\label{prompt:warning_diagnosis_user}
    \textbf{User Prompt for Warning Diagnosis}\newline
    Patient case: \newline
    \{case\_description\} \newline
    Briefly reason what the most dangerous diagnostic categories are for this presentation. Then pick exactly 3 diagnoses verbatim from the allowed list, one per distinct category. \newline
    Output ONLY a JSON list of exactly 3 items: \newline
    [
        \{``warning\_diagnosis'': ``<warning diagnosis>'', ``reason'': ``<$\leq$20 words>''\}, ...] \newline
    Output:
\end{prompt}

\begin{prompt}\label{prompt:diagnoses_review}
    \textbf{Prompt for Diagnoses Review}\newline
    Please analyze these medical diagnoses and cluster them into groups of similar conditions. You will be provided with a list of diagnoses, and you should group the similar diagnoses into different groups. \newline
        The group name should be a short description of the diagnosis, each diagnosis should be only in one group. Different group names mean different diagnoses. 
        You must return ONLY json objects, with the key: ``diagnosis'' (str) and ``group'' (str). The key is the orignal diagnosis and the value is the group name. 
        The output format must be:\newline
        [\{``diagnosis'': ``diagnosis 1'', ``group'': ``group 1''\},\newline
        \{``diagnosis'': ``diagnosis 2'', ``group'': ``group 2''\},\newline
        \{``diagnosis'': ``diagnosis 3'', ``group'': ``group 3''\}\}] \newline
        The provided diagnoses are: \{diagnosis\_list\} \newline
        Output:
\end{prompt}

\begin{prompt}\label{prompt:verification}
    \textbf{Prompt for Diagnosis Verification}\newline
    You are provided with a case description and a list of diagnoses. You task is to verify the diagnoses based on the confidence of the diagnosis for the case description.
        The case description is: \{case\_description\}. The diagnoses and reasonings are: \{diag\_reason\}.
        The verification score should be a number between 0 and 1, where 1 means the diagnosis is sufficiently supported by the case description, and 0 means the diagnosis is totally wrong.
        Please output the verified diagnoses in the following format:\newline
        [\{``diagnosis'': ``diagnosis 1'', ``verification\_score'': score1\},\newline
        \{``diagnosis'': ``diagnosis 2'', ``verification\_score'': score2\},\newline
        \{``diagnosis'': ``diagnosis 3'', ``verification\_score'': score3\},\newline
       \dots\}] \newline
        Output:

\end{prompt}

\begin{prompt}\label{prompt:query_search}
    \textbf{Prompt for Query Search}\newline
    Generate a PubMed search query to find reviews and clinical guidelines for the given disease or medical condition.
    Make sure the output can be used with the Entrez.esearch application programming interface (API). The search query should be as short as possible.\newline
    You can use these search features:\newline
    \textbullet{} Simple keyword search: ``covid vaccine''\newline
    \textbullet{} Field-specific search:\newline
    \textbullet{} Title search: [Title]\newline
    \textbullet{} Author search: [Author]\newline
    \textbullet{} MeSH terms: [MeSH Terms]\newline
    \textbullet{} Journal: [Journal]\newline
    \textbullet{} Date ranges: Add year or date range like ``2020:2024[Date - Publication]''\newline %
    \textbullet{} Combine terms with AND, OR, NOT\newline
    \textbullet{} Use quotation marks for exact phrases\newline
    User query: \{disease\}\newline
    Output the search query only, do not include any other text.

\end{prompt}

\begin{prompt}\label{prompt:reasoning}
    \textbf{Prompt for Single Diagnosis Reasoning}\newline
    You are provided with a case description, diagnosis result, and diagnosis references from multiple papers. Your task is to craft a concise, physician-style rationale for the diagnosis. 
        Keep the explanation focused no more than 30 words, or roughly 2 to 3 sentences. \newline
        Use the references to support the reasoning. Put the paper identifier (ID) in the end of the reasoning sentence if relevant.
        You must put all the used reference papers in the references list.  
        References should be in the following format:\newline
        [1]. Paper 1 title \newline
        [2]. Paper 2 title \newline
        ... \newline
        Here is the case description, diagnosis result, and diagnosis context from different papers:\newline
        Case Description: \{case\_description\}\newline
        Diagnosis: \{diagnosis\}\newline
        Diagnosis references: \{diagnosis\_context\}\newline
        Output:

\end{prompt}

\begin{prompt}\label{prompt:overall_reasoning}
    \textbf{Prompt for Overall Reasoning}\newline
    You are provided with a case description and multiple diagnoses. The diagnoses contain the possible diagnoses and the warning diagnoses that should not be missed for the case presentation. Each diagnosis has a corresponding reasoning and verification. \newline
            You task is to give the whole reasoning for all the diagnoses, considering the reasoning and verification for each diagnosis. \newline
            The reasoning should be concise and to the point, one of two summary sentences is present at the beginning. \newline
            For the case description \{case\_description\}. \newline
            The current diagnoses are \{diagnosis\_list\}. \newline
            The corresponding reasonings are \{reasoning\_list\}. \newline
            The corresponding verifications are \{verification\_list\}. \newline
            The warning diagnoses are \{warning\_diagnosis\_list\}. \newline
            Include all the diagnosis\_list and warning\_diagnosis\_list in the reasoning. Write the summary sentence of the reasoning at the beginning.
            Please include references based on the reasoning of each diagnosis. Put the paper ID in the end of the reasoning sentence if relevant. All the references should be present in the references list with ID in the beginning. Please cite the references in the following format: \newline
            [1]. Paper 1 title \newline
            [2]. Paper 2 title \newline
            ... \newline
            Please output the reasoning in the following json format, with the key ``reasoning'' and ``references'': \newline
            \{
                ``reasoning'': <str>,
                ``references'': <list>
            \} \newline
            Output:

\end{prompt}

\begin{prompt}\label{prompt:actions}
    \textbf{Prompt for Actions}\newline
    You are provided with a case description and a diagnosis. You task is to give what exact actions are needed to confirm that the diagnosis is correct.
        List the actions in a concise and to the point manner, and the actions should be specific and detailed.\newline
        For the case description \{case\_description\}. \newline
        The current diagnosis is \{diagnosis\}. \newline
        The previous verification is \{verification\}. \newline
        Please output the actions in the following format: \newline
        \{
            ``actions'': <str>
        \} \newline
        Output:

\end{prompt}

\begin{prompt}\label{prompt:management}
    \textbf{Prompt for Management Plan}\newline
    You are provided with a case description and multiple diagnoses. You task is to give the whole management plan considering all the diagnoses. \newline
        The management plan should be concise and to the point, one of two summary sentences is present at the beginning.
        For the case description \{case\_description\}. \newline
        The current diagnoses are \{diagnosis\_list\}. \newline
        The overall reasoning is \{overall\_reasoning\}. \newline
        The actions for each diagnosis are \{action\_list\}. \newline
        Please output the management plan in the following json format, with the key ``management\_plan'': \newline
        \{
            ``management\_plan'': <str>
        \} \newline
        Output:
\end{prompt}

\begin{prompt}\label{prompt:baseline_inference}
    \textbf{Prompt for Baseline Inference}\newline
    You are provided with a patient case. \newline 
Your task is to: \newline
1. List the top 3 diagnostic hypotheses for the patient case. \newline
2. List the top 3 must-not-miss diagnoses.\newline
3. Provide reasoning for each diagnosis.\newline
4. Provide the management plan for this patient case.\newline
Output should be in markdown format with the following 4 sections:\newline
1. Top 3 Diagnoses\newline
2. Top 3 Must-not-miss Diagnoses\newline
3. Reasoning\newline
4. Management\newline
The patient case is: \{case\_description\}\newline
    Output:
\end{prompt}

\begin{prompt}\label{prompt:evaluation}
    \textbf{Prompt for Evaluation}\newline
    Your task is to identify whether the provided predicted differential diagnosis is correct based on the true diagnosis. Carefully review the information and determine the correctness of the prediction. Please notice same diagnosis might be in different words. \newline
     Only return ``Y'' for yes or ``N'' for no, without any other words. \newline
     The true diagnosis is: \{true\_diagnosis\}\newline
     The predicted diagnosis is: \{predicted\_diagnosis\}\newline
     Output:
\end{prompt}

\section{Scaling with the Number of Specialist Agents}\label{app:num_agents_scaling}

\begin{figure*}[!t]
    \centering
    \includegraphics[width=\linewidth]{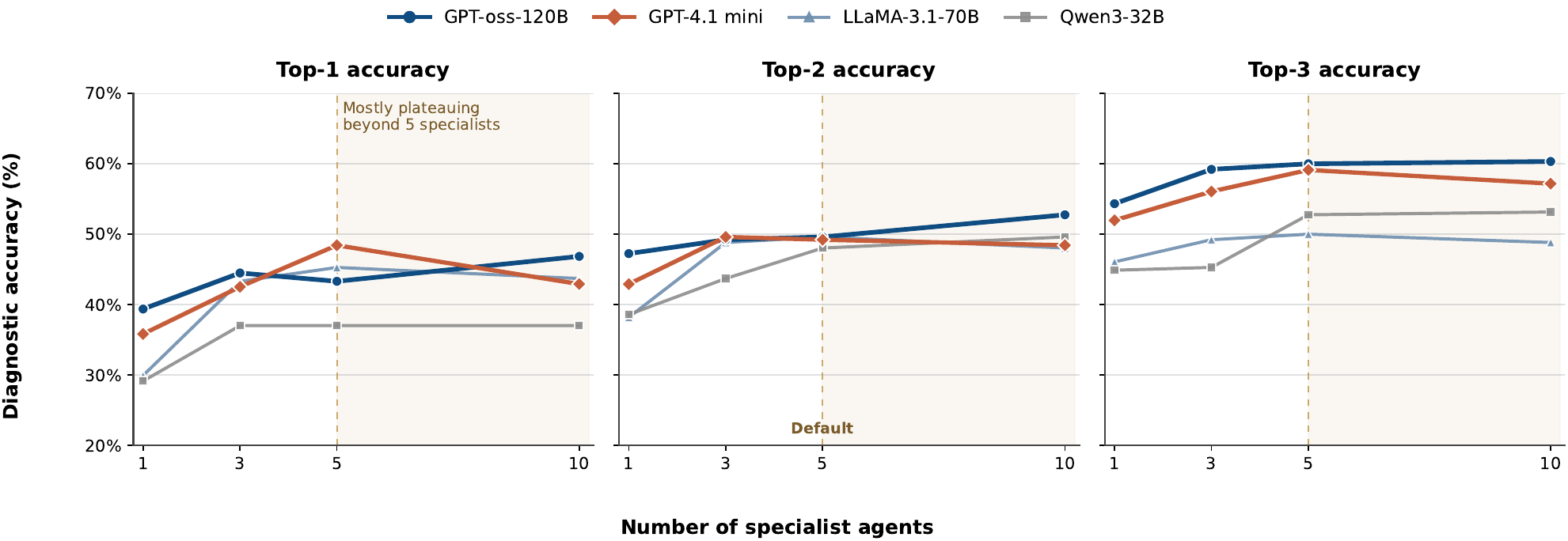}
    \caption{\textbf{Scaling with team size across model backbones.}
    Top-$1$, Top-$2$, and Top-$3$ diagnostic accuracy as a function of the number of specialist agents (1, 3, 5, and 10). Across most backbones, gains are largest when moving from a single specialist to a small team of 3--5 specialists, after which improvements become modest. GPT-oss-120B remains the strongest backbone across team sizes, whereas LLaMA-3.1-70B and GPT-4.1 mini show the steepest early Top-$1$ gains.}
    \label{fig:num_agents_scaling}
\end{figure*}

To define the optimal configuration for bedside deployment, we investigated how diagnostic performance scales with the number of specialized agents. We varied the specialist team size from $1$ to $10$ across multiple mid-sized and large-capacity backbone models (Figure~\ref{fig:num_agents_scaling}).

Diagnostic performance gains were most pronounced when increasing the number of agents from a single specialist to a compact ensemble of $3$ to $5$ specialists. Beyond this point, improvements began to plateau, with larger teams yielding diminishing returns. Marginal fluctuations in performance at higher team sizes likely reflect increased aggregation overhead and redundancy in agent reasoning, rather than foundation model limitations. This saturation behavior was consistent across model scales, suggesting that the default team size should balance hypothesis diversity against reasoning overlap. Based on these findings, we selected a compact five-specialist roster as the default AegisDx configuration for all subsequent experiments. This configuration provided a practical balance between diagnostic accuracy, inference efficiency, and rapid bedside response in time-pressured clinical settings.

\section{Interactive Web Application}\label{app:web_application}

\begin{figure}[!t]
    \centering
    \includegraphics[width=0.98\linewidth]{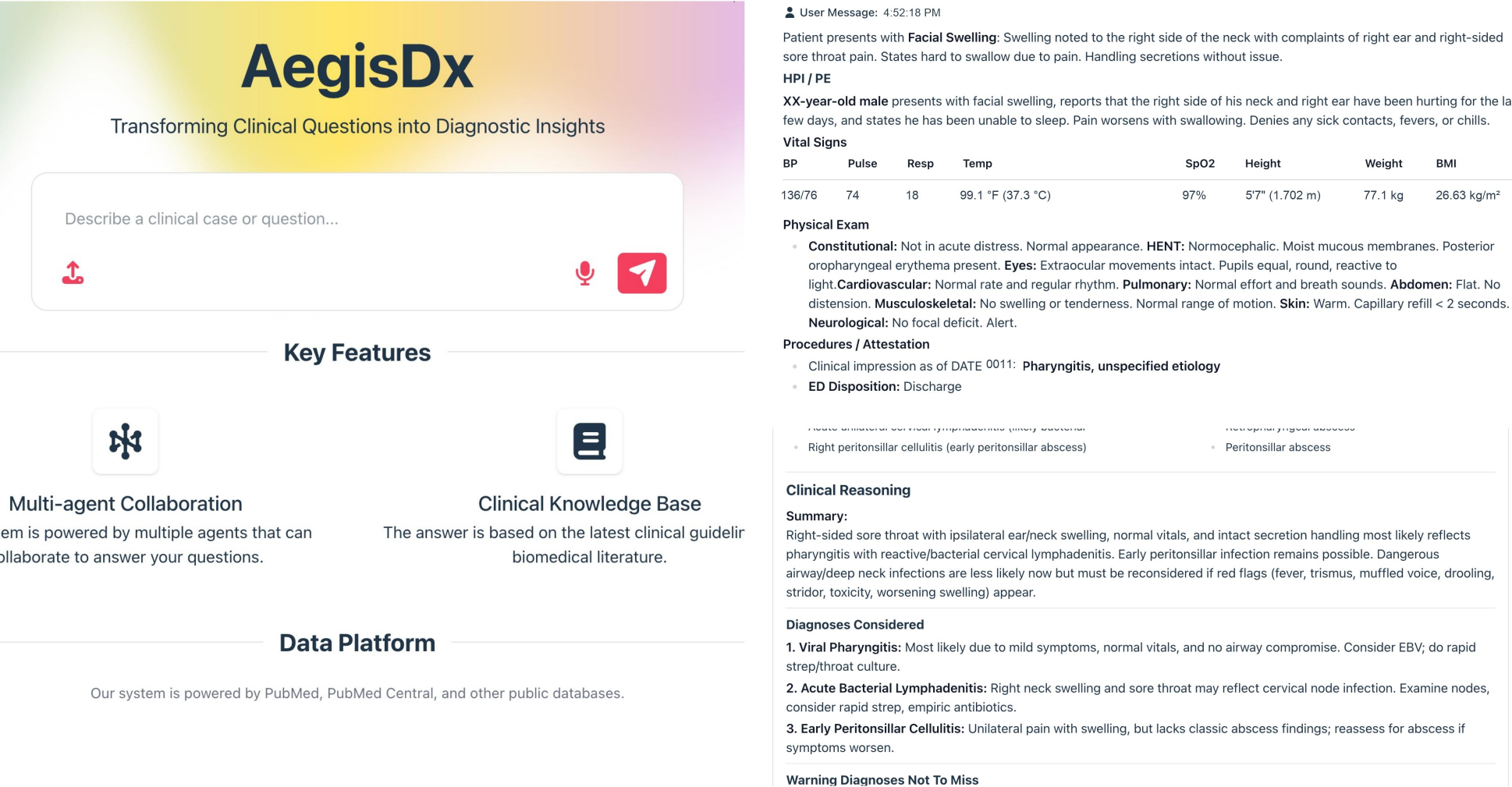}
    \caption{\textbf{Web application interface for AegisDx.} Left, case entry with free-text input and optional attachment/voice input. Right, output view presenting a structured clinical summary and AegisDx results, including reasoning, ranked diagnoses, and safety-oriented must-not-miss conditions.}
    \label{fig:web}
\end{figure}

To facilitate interactive use and clinical review, we developed a web application interface for \textbf{AegisDx}\footnote{https://clinicalnlp.org/AegisDx/}. As shown in Figure~\ref{fig:web}, the interface is designed to support the core clinician workflow in two steps.
First, users enter a case via free-text input (with optional attachment upload and voice input) along with key demographic context. Second, AegisDx generates clinician-readable results, including a concise reasoning summary, a ranked differential list, and safety-oriented must-not-miss conditions. In our deployment setting, AegisDx costs under \$1 per case and typically returns results in under $1$ minute, supporting timely clinical documentation and safety review.

\end{appendices}

\end{document}